\documentclass[lettersize,journal]{IEEEtran}
\usepackage{amsmath,amsfonts}
\usepackage{algorithmic}
\usepackage{algorithm}
\usepackage{array}
\usepackage{textcomp}
\usepackage{stfloats}
\usepackage{url}
\usepackage{verbatim}
\usepackage{graphicx}
\usepackage{cite}
\usepackage{multirow}
\usepackage{booktabs}
\usepackage{orcidlink}
\usepackage{makecell}
\usepackage{subcaption}

\begin{document}

\title{Cognition Transferring and Decoupling for Text-supervised Egocentric Semantic Segmentation}

\author{Zhaofeng Shi, Heqian Qiu,~\IEEEmembership{Member,~IEEE}, Lanxiao Wang,~\IEEEmembership{Student Member,~IEEE}, Fanman Meng,~\IEEEmembership{Member,~IEEE}
\\Qingbo Wu,~\IEEEmembership{Member,~IEEE}, Hongliang Li,~\IEEEmembership{Senior Member,~IEEE}

\thanks{This work was supported in part by the National Natural Science Foundation of China (No. U23A20286, No. 62301121 and No. 62271119), the China Postdoctoral Science Foundation under Grant 2023M740529, the Postdoctoral Fellowship Program (Grade B) of China Postdoctoral Science Foundation No.GZB20240120, and the Independent Research Project of Civil Aviation Flight Technology and Flight Safety Key Laboratory (FZ2022ZZ06).}
\thanks{The authors are with the School of Information and Communication Engineering, University of Electronic Science and Technology of China, Chengdu 611731, China (email: zfshi@std.uestc.edu.cn; hqqiu@uestc.edu.cn; lanxiao.wang@std.uestc.edu.cn; fmmeng@uestc.edu.cn; qbwu@uestc.edu.cn; hlli@uestc.edu.cn). Corresponding authors: Hongliang Li, Heqian Qiu, Lanxiao Wang.}

}

\markboth{IEEE Transactions on Circuits and Systems for Video Technology}%
{Shell \MakeLowercase{\textit{et al.}}: A Sample Article Using IEEEtran.cls for IEEE Journals}


\maketitle

\begin{abstract}
In this paper, we explore a novel Text-supervised Egocentic Semantic Segmentation (TESS) task that aims to assign pixel-level categories to egocentric images weakly supervised by texts from image-level labels. In this task with prospective potential, the egocentric scenes contain dense wearer-object relations and inter-object interference. However, most recent third-view methods leverage the frozen Contrastive Language-Image Pre-training (CLIP) model, which is pre-trained on the semantic-oriented third-view data and lapses in the egocentric view due to the ``relation insensitive" problem. Hence, we propose a Cognition Transferring and Decoupling Network (CTDN) that first learns the egocentric wearer-object relations via correlating the image and text. Besides, a Cognition Transferring Module (CTM) is developed to distill the cognitive knowledge from the large-scale pre-trained model to our model for recognizing egocentric objects with various semantics. Based on the transferred cognition, the Foreground-background Decoupling Module (FDM) disentangles the visual representations to explicitly discriminate the foreground and background regions to mitigate false activation areas caused by foreground-background interferential objects during egocentric relation learning. Extensive experiments on four TESS benchmarks demonstrate the effectiveness of our approach, which outperforms many recent related methods by a large margin. Code will be available at \href{https://github.com/ZhaofengSHI/CTDN}{https://github.com/ZhaofengSHI/CTDN}.
\end{abstract}

\begin{IEEEkeywords}
Text-supervised egocentric semantic segmentation, Cognition transferring, Foreground-background decoupling.
\end{IEEEkeywords}

\section{Introduction}
Understanding egocentric scenes, especially the camera wearer's hands and interacting objects, has become an essential computer vision task that helps artificial intelligence interpret the real world from a self-centered perspective. In recent years, researchers have pursued improving the granularity of egocentric parsing, i.e. egocentric image segmentation \cite{bambach2015lending,zhang2022fine,jia2022generative,cai2020generalizing}, which plays a pivotal role in various application scenarios: human-robot interaction \cite{ji2019survey,ji2020arbitrary,li2021survey,li2023proactive,dahiya2023survey,Qiu_2024_CVPR}, intelligent embodied devices \cite{zhang2020language,pu2023rules,nahavandi2022application,he2023learning,zhan2024enhancing}, and so on. Despite the impressive achievements, most existing egocentric segmentation models are trained in a fully-supervised manner, which requires time-consuming and labor-intensive pixel-level mask labeling. To alleviate this problem, some works \cite{zhang2022fine,lin2020ego2hands,gonzalez2020enhanced,yuan2021simple,ren2022adela} focus on data augmentation or automatic labeling strategies, whereas the diversity and accuracy of the generated annotations is limited. Li \textit{et al}. \cite{li2019supervised} resort to transfer information between different domains, while the data distribution gap causes inaccurate segmentation. Wu et al. \cite{wu2023continual} propose an awesome continual image segmentation method that only requires annotations for the newly emerged classes during the incremental learning stages. Many researchers \cite{wei2016stc,kolesnikov2016seed,ahn2018learning,xu2022multi,Xie_2022_CVPR,ru2022learning,ru2023token,xu2023mctformer+} attempt to leverage readily available annotation (i.e. image-level classification labels) to generate and refine the initial class-activation maps (CAM), which rely only on visual modality and lack the suppression of the related background regions \cite{xie2022clims}, resulting in suboptimal pseudo mask quality. We observe that the texts can provide distinctive descriptions for each unique object in the egocentric scenes, and they are convenient to acquire from the image-level labels. Moreover, powerful image-text pre-trained models \cite{radford2021learning,jia2021scaling} have been well-studied that bridge worldwide visual and textual cognition, which elicits a question: \textit{Can texts serve as the supervision for the egocentric semantic segmentation?}
\begin{figure}[!t]
\centering
\includegraphics[width=0.90\linewidth]{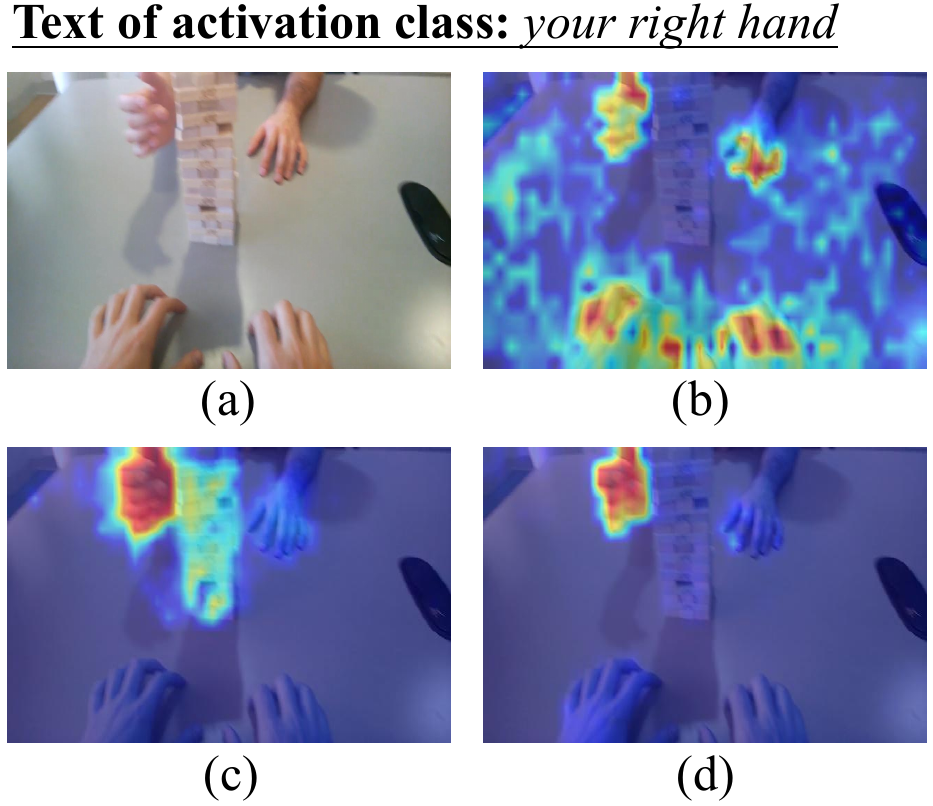}
\caption{Visualizations of different models according to the given class text ``\textit{your right hand}". (a): the input egocentric image. (b): the class-activation map (CAM) of the original CLIP model. (c): the CAM of the finetuned CLIP model. (d): the CAM of our CTDN model.}
\label{fig:1}
\end{figure}

To support this research, in this paper, we make the first exploration of a novel text-supervised egocentric semantic segmentation (TESS) task, which aims to conduct semantic segmentation on egocentric images based on weak text supervision from image-level labels. The fine-grained parsing of crucial camera wearer's hands and interacting objects in egocentric scenarios is important, and it is a fundamental task in many emerging fields such as embodied AI \cite{nilsson2021embodied}. However, the mask labeling of egocentric images is difficult because it requires identifying active objects based on hand-object contact and outlining boundaries of multiple interacting objects in local regions. In practice, a skilled labeler takes an average labeling time of up to above 50 seconds per object \cite{darkhalil2022epic}, which obstructs large-scale egocentric data annotation. The proposed TESS task is motivated by leveraging texts that are easily accessible and descriptive to egocentric objects as supervision information to alleviate the burden of labeling and cost-effectively training egocentric semantic segmentation models. Recently, some third-view weakly-supervised semantic segmentation (WSSS) methods \cite{xie2022clims,lin2023clip} introduce the text modality via the Contrastive Language-Image Pre-training (CLIP) \cite{radford2021learning} model and achieve remarkable performance in common third-view scenes. Nevertheless, it is difficult for such CLIP-based third-view WSSS methods to cover the egocentric scenes due to the CLIP's ``relation insensitive" problem. Specifically, the egocentric images contain dense relations between the camera wearer and objects of interest such as subordination (\textit{my, your}) and location (\textit{left, right}), while the CLIP model is pre-trained on the semantic-oriented but relation sparse third-view data. Therefore, the CLIP mainly concentrates on the objects' semantic information rather than distinguishing objects with identical semantics but different wearer-object relations such as \textit{my or your hands}, results in column (b) of Fig. \ref{fig:1}: all the hands are highlighted with severe noises. An intuitive way to address the problem is to directly finetune the CLIP to distinguish foreground objects with dense relations. However, such finetune-based learning causes degradation of the CLIP's inherent capability of discriminating foreground and background objects with various semantics, which causes the false activation of the irrelevant background objects. For example, the interested \textit{hand} and uninterested \textit{jenga} frequently contact and tightly adhere in egocentric images, which causes inter-object interference and misleads the model to also highlight the unconcerned background areas as in column (c) of Fig. \ref{fig:1}, which leads to misclassification regions in the pseudo masks.

To address the above problem, we propose a Cognition Transferring and Decoupling Network (CTDN) for the TESS task. Specifically, ``cognition" refers to the information of the fore/background contents in the egocentric view extracted by the on-the-shelf pre-trained model. Unlike other CLIP-based methods that distill from visual representations \cite{gu2021open,wang2022clip,wu2023clipself,chen2023exploring} or class texts-based logits \cite{wu2023tinyclip,rasheed2023fine} to enhance the student model's third-view open set capability, our CTDN not only learns egocentric relations, but also transfers the knowledge based on an extra well-designed cognition set to prevent the degradation of the fore/background object recognition ability in egocentric scenes, which facilitates the representation decoupling to explicitly discriminate the foreground and background regions. In detail, we first learn the egocentric wearer-object relations through multi-label classification by measuring the image-text cross-modal similarity. Then, we propose a Cognition Transferring Module (CTM), which leverages the frozen CLIP visual and text encoders to calculate the semantic-level probability distribution responses between the egocentric image and cognition texts, which serves as the constraint of our egocentric encoders to facilitate its recognition of the foreground and background semantics. Besides, feature-level information distillation is also adopted to maintain the representation stability. Based on the transferred cognition, we adopt the Foreground-background Decoupling Module (FDM) to calculate the prototype vectors of foreground and background, and further categorize and contrastively disentangle the visual representations. Thereby, the model can accurately activate the object based on the given class text as shown in column (d) of Fig. \ref{fig:1}, and generate precise pseudo masks.

In summary, the major contributions of the paper can be concluded as follows:
\begin{itemize}
    \item To the best of our knowledge, it is the first exploration of the Text-supervised Egocentric Semantic Segmentation (TESS) task. In addition, we propose a Cognition Transferring and Decoupling Network (CTDN) to learn egocentric wearer-object relations and mitigate the false activation problem caused by the foreground-background interferential objects.
    \item We develop a Cognition Transferring Module (CTM) to transfer cognitive knowledge from the large-scale pre-trained model to our egocentric model for recognizing frequent interfering objects. And a Foreground-background Decoupling Module (FDM) is proposed to contrastively disentangle the visual representations based on the transferred cognition to explicitly discriminate the foreground and background regions.
    \item Extensive experiments on four text-supervised egocentric semantic segmentation benchmarks demonstrate the effectiveness of the proposed method, which outperforms other methods from the related task by a large margin.
\end{itemize}

\section{Related Work}
\subsection{Egocentric Hand-Object Interaction Understanding}
Despite the extensive research that has been conducted on the third-view human action recognition \cite{kong2022human,ji20123d,yao2019review,xing2023svformer,cao2023vs},  there is less attention to the egocentric scenes. Egocentric data \cite{grauman2022ego4d,damen2018scaling,sigurdsson2018charades} is typically collected by the head-mounted device and mainly concentrates on the actions consisting of hands and interaction objects in the wearer's view. Egocentric Hand-Object Interaction (Ego-HOI) understanding is essential to various applications such as augmented reality \cite{arena2022overview}. To facilitate Ego-HOI understanding, researchers have proposed many corresponding datasets \cite{zhu2023egoobjects,liu2022hoi4d}, and CNN-based \cite{lu2019learning} and RNN-based \cite{sudhakaran2019lsta,furnari2020rolling} frameworks. In recent years, many methods \cite{cho2022transformer,xu2023egopca} based on Transformers \cite{vaswani2017attention} have been proposed and achieved remarkable performance. In addition, researchers also make other attempts such as cross-modal pre-training \cite{lin2022egocentric,pramanick2023egovlpv2}, multi-modal understanding \cite{radevski2023multimodal,gong2023mmg}, open-vocabulary action recognition \cite{chatterjee2023opening}, prompt learning \cite{yu2023efficient,xu2023pov} and continual learning \cite{xu2023towards}, which make great contributions to this field.

Recently, researchers have become increasingly interested in the challenging fine-grained Ego-HOI understanding, which means bounding box-level or pixel-level extraction of hands and interacting objects in the egocentric view rather than merely image-level classification. To support the following research, Zhu \textit{et al}. \cite{zhu2023egoobjects} propose an Ego-HOI understanding dataset, and Kurita \textit{et al}. \cite{kurita2023refego} propose an egocentric referring comprehension dataset. Bambach \textit{et al}. \cite{bambach2015lending} and Zhang \textit{et al}. \cite{zhang2022fine} propose an egocentric dataset with pixel-level annotations. There are also many egocentric localization \cite{wang2023ego,huang2023egocentric,wu2023localizing} and segmentation \cite{cai2020generalizing,gonzalez2022real,yemisi2023fine} methods have been proposed. Different from the above works, our method aims to perform egocentric semantic segmentation of hands and interacting objects supervised only on the texts from the image-level label, which is under-explored.

\begin{figure*}[!t]
\centering
\includegraphics[width=1.0\linewidth]{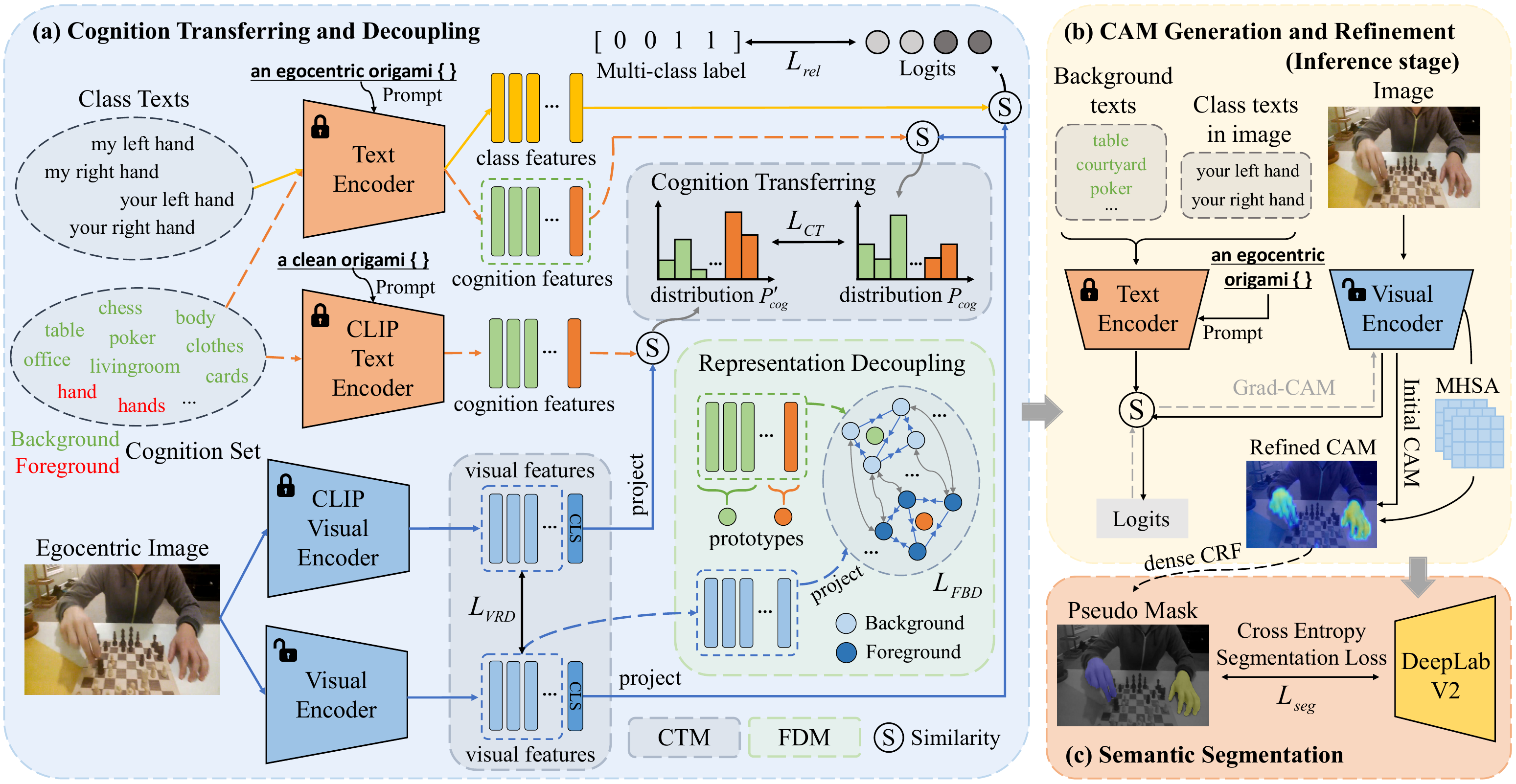}
\caption{The overview of three-stage CTDN. In the first stage, given the egocentric image, class texts, and cognition set, we extract the corresponding features and first learn the egocentric relations. In addition, the Cognition Transferring Module (CTM) and Foreground-background Decoupling Module (FDM) are developed to transfer the cognitive knowledge into our egocentric model and disentangle the foreground and background representations. In the second stage, we perform backpropagation to obtain the initial CAM and refine it by MHSA. In the third stage, the pseudo masks are generated and a DeepLab V2 model \cite{chen2017deeplab} is adopted for the egocentric semantic segmentation.}
\label{fig:2}
\end{figure*}

\subsection{Semantic Segmentation using Less Supervision}
Many methods have been proposed to use less supervision to reduce the burden of pixel-level labeling for semantic segmentation. Some few-shot \cite{lu2021simpler,liu2020part,yang2020prototype} or zero-shot \cite{bucher2019zero,li2020consistent,ding2022decoupling} learning methods aim to perform semantic segmentation with few or without pixel-level masks for novel categories. Another research direction is weakly-supervised semantic segmentation (WSSS), which typically aims to generate pseudo masks using image-level classification labels. Early WSSS methods \cite{ahn2018learning,kolesnikov2016seed,huang2018weakly} utilize the class-activation map (CAM) \cite{he2016deep} to highlight the most discriminative region, which causes incomplete object areas. In the following studies, some methods \cite{li2021pseudo,lee2021anti,lee2022threshold,su2021context,zhang2020reliability,zhang2021complementary} attempts to complement the initial CAM. Some other methods \cite{zhang2021adaptive,ru2022learning,wang2020weakly,ru2023token} attempt to learn pixel-wise or patch-wise affinity for CAM refinement. Some researchers aim to leverage auxiliary information such as the saliency map \cite{lee2021railroad,wang2020self,yao2021non}, cross-image information \cite{sun2020mining,liu2021cross,ru2022weakly}, and unsupervised models \cite{jo2023mars} to remove biases of the WSSS framework. In recent years, due to the advantage of the Vision Transformer \cite{vaswani2017attention,dosovitskiy2020image} for modeling long-range dependencies, lots of researchers \cite{xu2022multi,xu2023mctformer+} have used it in the WSSS frameworks. CLIMS \cite{xie2022clims} and CLIP-ES \cite{lin2023clip} attempt to utilize CLIP \cite{radford2021learning} in the WSSS task to match the image and text inputs and generate accurate object regions. In addition to WSSS, various works based on data augmentation \cite{yuan2021simple,zhang2022fine}, domain adaptation \cite{li2020content,lv2020cross}, and automatic labeling \cite{ren2022adela,qiao2023human} are also developed to alleviate the pixel-level annotation problem. Despite the above remarkable achievements in third-view scenarios, the research towards weakly-supervised semantic segmentation in the egocentric view is limited.

\section{Method}
The overall architecture of our Cognition Transferring and Decoupling Network (CTDN) is illustrated in Fig. \ref{fig:2}. The framework is composed of three stages. In the cognition transferring and decoupling stage, an egocentric image $I\in {{\mathbb{R}}^{H\times W\times 3}}$ and the class texts ${{T}_{cls}}$ are first leveraged to learn the egocentric wearer-object relations. Next, we additionally set the foreground and background cognition set $\mathcal{C}$ and propose the Cognition Transferring Module (CTM) to transfer cognitive knowledge from the frozen CLIP into our model. Then, the Foreground-background Decoupling Module (FDM) disentangles the visual representations to explicitly discriminate the foreground and background regions to mitigate false activation. In the CAM generation and refinement stage, we use Grad-CAM \cite{selvaraju2017grad} to get the initial class-activation map (CAM) and utilize the Transformer's multi-head self-attention (MHSA) weights for CAM refinement. In the semantic segmentation stage, we generate the pseudo masks based on the refined CAM and perform egocentric semantic segmentation.

The CTDN is under egocentric settings since its key components are intended to overcome the egocentric-specific challenges (i.e. dense wearer-object relations and complex foreground-background object interference). Specifically, the CTDN learns egocentric relations by correlating the egocentric images focusing on hand-object interactions and class texts containing phrases describing egocentric wearer-object relations with the egocentric prompt template. Moreover, the proposed CTM performs knowledge distillation based on the cognition set consisting of frequently interfering objects or contents in egocentric contexts, facilitating the FDM to disentangle the visual representations of interferential foreground and background objects in egocentric scenes. 

Note that the ``text-supervised" means that the proposed CTDN framework only utilizes the annotations of image-level labels and the corresponding class label texts to supervise the egocentric model training in the first Cognition Transferring and Decoupling stage. The trained model is used to accurately generate the initial CAM and pseudo masks, which is the main objective of various weakly-supervised segmentation tasks \cite{Li_2021_ICCV,liu2023referring}. The final semantic segmentation stage follows the common setting in previous works \cite{xie2022clims,xu2023mctformer+} to train a universal segmentation model such as DeepLab, which serves as a refinement process for the generated pseudo masks.

\subsection{Egocentric Relation Learning}
Despite the CLIP's capability to bridge third-view visual and textual semantic objects, it cannot adapt to egocentric views with relations such as subordination and location. To solve the egocentric relation learning, the proposed model correlates the relation-abundant egocentric images and their corresponding class texts to learn dense egocentric relations. In detail, the model calculates the cross-modal similarities between the extracted visual and textual features as predicted logits and correlates them via a multi-label classification task, which facilitates understanding the correspondence between multiple egocentric visual objects with complex interactions and class texts with egocentric wearer-object relations.

Given the egocentric image $I$ and class texts ${{T}_{cls}}=\{{{t}_{1}},{{t}_{2}},\cdots {{t}_{T}}\}$ with relation descriptions, we adopt a visual encoder ${\mathcal{E}_{V}}(\cdot )$ and a text encoder ${\mathcal{E}_{T}}(\cdot )$ with Transformer-based structures \cite{vaswani2017attention,dosovitskiy2020image} to extract their features, as follows:
\begin{equation}
{{F}_{V}}={\mathcal{E}_{V}}(I)\text{    }\text{    }\text{    }\text{    }{{F}_{T}}={\mathcal{E}_{T}}({p}({T}_{cls}))
\end{equation}
where ${{F}_{V}}\in {{\mathbb{R}}^{{{N}_{v}}\times {{C}_{v}}}}$ and ${{F}_{T}}\in {{\mathbb{R}}^{{{N}_{t}}\times {C}}}$ denotes the extracted visual features and class features. ${N}_{v}$, ${N}_{t}$ are the respective number of visual tokens and classes. ${C}_{v}$, ${C}$ represents the visual/text feature dimensions, $p$ denotes the prompt template.

We correlate the egocentric images and texts by measuring the cross-modal similarities and perform multi-label classification for coarsely learning the egocentric relations. The ground-truth classification label is denoted as ${{Y}_{cls}}\in {{\{0,1\}}^{1\times {{N}_{t}}}}$. The logits ${{\tilde{Y}}_{cls}}=\{{{\tilde{y}}_{cls,i}}\}_{i=1}^{{{N}_{t}}}$ can be predicted by measuring the similarities between the projected visual class token ${{f}_{cls}}\leftarrow {{F}_{V}}(0)$ and class text feature vectors ${{F}_{T}}=\{{{f}_{t,i}}\}_{i=1}^{{{N}_{t}}}$, which is formulated as follow:
\begin{equation}
\label{eq:simi}
{{\tilde{y}}_{cls,i}}=\frac{{{f}_{cls}}\cdot {{({{f}_{t,i}})}^{T}}}{||{{f}_{cls}}|{{|}_{2}}\cdot ||{{f}_{t,i}}|{{|}_{2}}}\cdot \mu 
\end{equation}
where $\mu$ is a learnable scale factor. Finally, we train our model to learn the relations using a binary cross-entropy (BCE) loss, which can be denoted as ${{L}_{rel}}$.

\subsection{Cognition Transferring Module}
Since the foreground-background interferential objects like \textit{hands} and \textit{jenga}, \textit{hands} and \textit{table}, frequently contact and interfere with each other in a single egocentric image, direct finetuning causes false activation of the background areas and leads to misclassification in the generated pseudo mask. Therefore, as shown in the purple dashed boxes of Fig. \ref{fig:2}, we propose a Cognition Transferring Module (CTM), which aims to transfer the cognitive knowledge of the large-scale pre-trained model into our model to enable the recognition of the semantics of fore/background objects, and guides the subsequent representation decoupling.

Inspired by the idea of Lin \textit{et al}. \cite{lin2023clip}, we design a cognition set $\mathcal{C}=\{{{c}_{1}},{{c}_{2}},\cdots {{c}_{{N}_{cog}}}\}$ that contains words or phrases of the frequently interfering objects or contents in the foreground and background. And we adopt ViT-based CLIP visual and text encoders ${\mathcal{E}'_{V}}(\cdot )$ and ${\mathcal{E}'_{T}}(\cdot )$ to extract the visual and cognition features, respectively. Additionally, because the prompt ensemble strategies do not perform well on the weakly-supervised semantic segmentation \cite{lin2023clip}, we design unique manual prompt templates for our egocentric text encoder ${\mathcal{E}_{T}}(\cdot )$ and the CLIP text encoder ${\mathcal{E}'_{T}}(\cdot )$ as the semantic contextual guidance. Specifically, the prompt ${p}$ for our egocentric text encoder is `\textit{an egocentric origami \{\}.}', and the prompt ${\tilde{p}}$ is set to `\textit{a clean origami \{\}.}' for the original CLIP text encoder. Then, we extract the visual and text features of the egocentric image and the cognition set, which are formulated as follows:
\begin{equation}
{{F}_{V}}={\mathcal{E}_{V}}(I)\text{    }\text{    }\text{    }\text{    }{{{F}}'_{V}}={{\mathcal{E}}'_{V}}(I)
\end{equation}
\begin{equation}
{{F}_{C}}={\mathcal{E}_{T}}({p}(\mathcal{C}))\text{    }\text{    }\text{    }\text{    }{{{F}}'_{C}}={{\mathcal{E}}'_{T}}(\tilde{p}(\mathcal{C}))
\end{equation}
where ${{F}_{V}},{{F}'_{V}}\in {{\mathbb{R}}^{{{N}_{v}}\times {{C}_{v}}}}$ are extracted visual features, and ${{F}_{C}}, {{F}'_{C}}\in {{\mathbb{R}}^{{{N}_{cog}}\times C}}$ are cognition textual features. In practice, to accurately extract class features ${{F}_{T}}$ and cognition features ${{F}_{C}}$ including fore/background features, the text encoder ${\mathcal{E}_{T}}(\cdot )$ is frozen during training and initialized with the parameters of the CLIP text encoder ${\mathcal{E}'_{T}}(\cdot )$, which is pre-trained on world-scale textual corpora covering the words or phrases in the class and cognition sets and is capable of extracting robust textual features. Moreover, the visual encoder ${\mathcal{E}_{V}}(\cdot )$ is trainable to extract the egocentric visual features, which are correlated with the extracted textual features to facilitate the learning of the egocentric relations and representations.

Next, we obtain the respective logits of our egocentric encoders and the CLIP encoders by calculating the cross-modal similarities between the projected visual class token and cognition features as in Eq. \ref{eq:simi}, which can be denoted as ${{Y}_{cog}}=\{{{y}_{cog,i}}\}_{i=1}^{{{N}_{cog}}}$ and ${{Y}'_{cog}}=\{{{y}'_{cog,i}}\}_{i=1}^{{{N}_{cog}}}$, respectively. The probability distribution responses ${P}_{cog}$, ${P}'_{cog}$ can be computed via \textit{Softmax} normalization. Finally, we conduct the semantic-level cognition transferring from the third-view to egocentric through the Kullback-Leibler (KL) divergence constraint, and the cognition transferring loss ${{L}_{CT}}$ can be formulated as follows:
\begin{equation}
{{L}_{CT}}=KL({{P}_{cog}}||{{{P}'}_{cog}})=\sum\limits_{i}{{{P}_{cog}}(i)\log \frac{{{P}_{cog}}(i)}{{{{P}'}_{cog}}(i)}}  
\end{equation}
Furthermore, for further transferring the cognitive knowledge from the feature level to maintain the stability of the learned visual representations, we adopt a visual representation distillation loss ${L}_{VRD}$, as follows:
\begin{equation}
{{L}_{VRD}}=MSE(\psi (\frac{1}{{{N}_{v}}-1}\sum\limits_{i\ne 0}{{{F}_{V}}(i)}),{\psi }'(\frac{1}{{{N}_{v}}-1}\sum\limits_{i\ne 0}{{{F}'_{V}}(i)}))
\end{equation}
where $\psi (\cdot )$, ${\psi}' (\cdot )$ means the project layers of the respective encoders. The transferred cognition enables our egocentric encoders to recognize fore/background objects, which facilitates the subsequent visual representation decoupling.

Note that the difference between the text encoder ${\mathcal{E}_{T}}(\cdot )$ and the CLIP text encoder ${\mathcal{E}'_{T}}(\cdot )$ lies in their inputs and functions. Specifically, the text encoder takes the class texts and cognition set with the egocentric prompt ${p}$ as inputs. On the one hand, it extracts cognition features to calculate visual-textual similarity-based logits as responses of the student model for knowledge transferring. On the other hand, the class features are extracted to correlate with the visual features from a trainable visual encoder to learn egocentric relations. In contrast, the CLIP text encoder takes the cognition set with a general prompt ${\tilde{p}}$ as input, which is used to calculate logits with the visual features from the frozen CLIP visual encoder as the responses of the teacher model as guidance during training.

\subsection{Foreground-background Decoupling Module}

\begin{figure}[!t]
\centering
\includegraphics[width=1.0\linewidth]{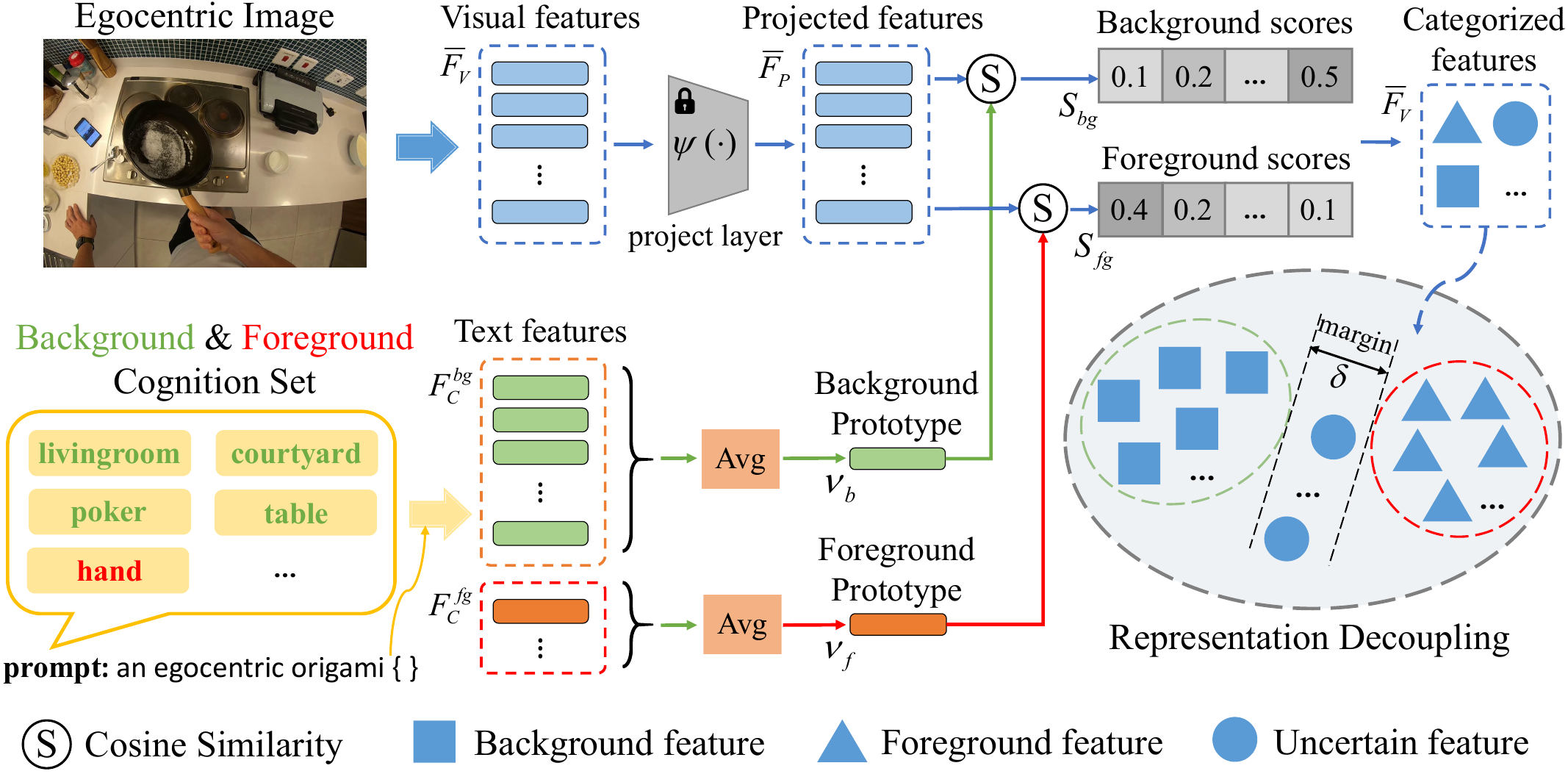}
\caption{Schematic of FDM. We first extract the cognition features and compute the prototypes of foreground and background. Then, we calculate the similarities between the projected visual features and prototypes to get the respective scores, based on which we conduct representation decoupling for the fore/background features.}
\label{fig:3}
\end{figure}

The transferred cognition facilitates our model to semantically and representationally recognize the foreground and background objects. Moreover, as shown in Fig. \ref{fig:3}, we propose a Foreground-background Decoupling Module (FDM) to contrastively disentangle the foreground and background visual representations. The motivation is that objects in the egocentric scenes frequently contact and tightly adhere with each other, which causes complex foreground-background object interference and false activation towards the unconcerned background regions. To address the above challenge, the FDM performs contrastive training based on the extracted foreground/background cognition features to explicitly disentangle the visual representations, which facilitates discriminating the foreground and background regions in egocentric scenes and mitigating false activation areas in the generated pseudo masks. Different from standard contrastive learning approaches that treat images with different views but identical contents \cite{chen2020simple} or matched image-text pairs \cite{radford2021learning} within a batch as positive samples and other cases as negative samples to learn coarse image-level representations or cross-modal correspondence. The proposed FDM categorizes the visual representations into foreground and background sets by calculating the cross-modal similarities with the cognition-based prototypes. Then, it considers visual representations in the same foreground/background set as positive samples and vice versa to conduct fine-grained pixel-level contrastive learning.

We further split the cognition set $\mathcal{C}$ into foreground set ${{\mathcal{C}}_{fg}}=\{c_{1}^{f},c_{2}^{f},\cdots ,c_{{{N}_{fg}}}^{f}\}$ and background set ${{\mathcal{C}}_{bg}}=\{c_{1}^{b},c_{2}^{b},\cdots ,c_{{{N}_{bg}}}^{b}\}$, where ${{N}_{fg}}$ and ${{N}_{bg}}$ mean the number of foreground and background texts in the cognition set, respectively. And their corresponding features can be denoted as $F_{C}^{fg}=\{f_{c,i}^{fg}\}_{i=1}^{{{N}_{fg}}}\in {{\mathbb{R}}^{{{N}_{fg}}\times C}}$ and $F_{C}^{bg}=\{f_{c,j}^{bg}\}_{j=1}^{{{N}_{bg}}}\in {{\mathbb{R}}^{{{N}_{bg}}\times C}}$. Then, we compute the average of the corresponding features to obtain the foreground and background prototype vectors ${{\nu }_{f}}$ and ${{\nu }_{b}}$. For the extracted visual features ${{F}_{V}}\in {{\mathbb{R}}^{{{N}_{v}}\times {{C}_{v}}}}$, we remove the special class token and the remaining visual tokens can be denoted as ${{\bar{F}}_{V}}=\{{{\bar{f}}_{v,i}}\}_{i=1}^{{{N}_{v}}-1}\in {{\mathbb{R}}^{({{N}_{v}}-1)\times {{C}_{v}}}}$. Since the visual tokens have been transferred with cognitive knowledge, they contain high-level modality-unified semantic information. Therefore, we feed the visual tokens into the frozen cross-modal project layer, as follows:
\begin{equation}
{{\bar{F}}_{P}}={{\psi }_{frozen}}({{\bar{F}}_{V}})
\end{equation}
where ${{\bar{F}}_{P}}=\{{{\bar{f}}_{p,i}}\}_{i=1}^{{{N}_{v}}-1}\in {{\mathbb{R}}^{({{N}_{v}}-1)\times C}}$ denotes the projected visual features. Next, we compute the cosine similarity between the projected visual features ${{\bar{F}}_{P}}$ and the prototype vectors ${{\nu }_{f}}$, ${{\nu }_{b}}$ respectively to obtain the foreground and background scores, which are represented as ${{S}_{fg}}=\{s_{1}^{f},s_{2}^{f},\cdots s_{{{N}_{v}}-1}^{f}\}$ and ${{S}_{bg}}=\{s_{1}^{b},s_{2}^{b},\cdots s_{{{N}_{v}}-1}^{b}\}$. Based on the calculated scores, the visual features can be categorized into foreground $\mathcal{{P}}_{f}$ and background $\mathcal{{P}}_{b}$. In addition, we introduce a margin hyperparameter $\delta$ and an uncertain $\mathcal{{P}}_{u}$ feature set, the features of which are excluded in the following contrastive learning. The visual feature categorizing process can be formulated as follows:
\begin{equation}
    \begin{aligned}
\left\{ 
\begin{matrix}
   {{{\bar{f}}}_{v,i}}\in {{\mathcal{P}}_{f}}\text{   }\text{   }\text{   }\text{   }\text{   }\text{   }\text{   }\text{   }\text{   }\text{   }\text{                if  }s_{i}^{f}-s_{i}^{b}>\delta   \\
   {{{\bar{f}}}_{v,i}}\in {{\mathcal{P}}_{u}}\text{   }\text{   }\text{   }\text{   }\text{   }\text{   }\text{   }\text{   }\text{                if  }|s_{i}^{f}-s_{i}^{b}|\le \delta   \\
   {{{\bar{f}}}_{v,i}}\in {{\mathcal{P}}_{b}}\text{   }\text{   }\text{   }\text{   }\text{   }\text{   }\text{   }\text{   }\text{                if  }s_{i}^{f}-s_{i}^{b}<-\delta   \\
\end{matrix}
\right.
    \end{aligned}
\end{equation}
Finally, we perform contrastive decoupling of visual representations. Specifically, if two representations both belong to the foreground or background category, they are considered positive samples. Otherwise, they are treated as negative samples. And the representations in the uncertain set are excluded. The foreground-background decoupling loss ${L}_{FBD}$ can be calculated as follows:
\begin{multline}
{{L}_{FBD}}=\frac{0.25}{N_{f}^{+}}\sum\limits_{{{{\bar{f}}}_{v,i}},{{{\bar{f}}}_{v,j}}\in {{\mathcal{P}}_{f}}}{(1-\vartheta ({{{\bar{f}}}_{v,i}},{{{\bar{f}}}_{v,j}}))}\\+\frac{0.25}{N_{b}^{+}}\sum\limits_{{{{\bar{f}}}_{v,i}},{{{\bar{f}}}_{v,j}}\in {{\mathcal{P}}_{b}}}{(1-\vartheta ({{{\bar{f}}}_{v,i}},{{{\bar{f}}}_{v,j}}))}\\+\frac{0.5}{{{N}^{-}}}\sum\limits_{{{{\bar{f}}}_{v,i}}\in {{\mathcal{P}}_{f}},{{{\bar{f}}}_{v,j}}\in {{\mathcal{P}}_{b}}}{\vartheta ({{{\bar{f}}}_{v,i}},{{{\bar{f}}}_{v,j}})}
\end{multline}
where ${N_{f}^{+}}$, ${N_{b}^{+}}$, ${{N}^{-}}$ denotes the number of foreground positive pairs, background positive pairs, negative pairs, respectively. And $\vartheta$ means the absolute cosine similarity. Afterward, visual representations with the same fore/background semantics are pulled closer, while representations with different semantics are decoupled. The overall loss in the first stage can be calculated as follow:
\begin{equation}
L={{L}_{rel}}+{{\lambda }_{1}}{{L}_{CT}}+{{\lambda }_{2}}{{L}_{VRD}}+{{\lambda }_{3}}{{L}_{FBD}}
\end{equation}
where ${{\lambda }_{1}}$, ${{\lambda }_{2}}$, and ${{\lambda }_{3}}$ are hyperparameters to balance the multiple loss items.

\subsection{Pseudo Mask Generation and Segmentation}
Before generating the final pseudo mask, we first generate and refine the class-activation map (CAM) as shown in Fig. \ref{fig:2}. In detail, we first extract the features of the image $I$ and the appeared classes ${{T}'_{cls}}$. Moreover, we add the background cognition texts as inputs to facilitate the suppression of the background regions. Then, we obtain the logits ${{\tilde{Y}}'}=\{{{\tilde{y}}'_{i}}\}_{i=1}^{{{N}'}}$ by measuring the similarities. To obtain the initial CAM, we utilize Grad-CAM \cite{selvaraju2017grad}, which performs backpropagation and calculates the weighted sum of the activation as follows:
\begin{equation}
\omega _{z}^{c}=\frac{1}{H\times W}\sum\limits_{x}{\sum\limits_{y}{\frac{\partial {{{\tilde{{y}}}}'_{c}}}{\partial A_{xy}^{z}}}}
\end{equation}
\begin{equation}
{{M}^{c}_{xy}}=ReLU(\sum\limits_{z}{\omega _{z}^{c}A_{xy}^{z}})
\end{equation}
where ${{{\tilde{{y}}}}'_{c}}$ denotes the $c$-$th$ class logit, $A$ denotes activation feature map, $\omega$ is the class-activated weights, and ${M}^{c}$ denotes the initial CAM of class $c$. We extract the multi-head self-attention (MHSA) weights ${{F}_{MHSA}}\in {{\mathbb{R}}^{hw\times hw}}$ to refine the initial CAM $M\in {{\mathbb{R}}^{h\times w}}$, which can be formulated as follows:
\begin{equation}
{{\Phi }_{Aff}}=\frac{\sigma ({{F}_{MHSA}})+\sigma {{({{F}_{MHSA}})}^{T}}}{2}
\end{equation}
\begin{equation}
{{\tilde{M}}^{c}}={{B}^{c}}\odot {{\Phi }_{Aff}}\cdot vec({{M}^{c}})
\end{equation}
where $\sigma$ denotes \textit{Sinkhorn} normalization function, ${{B}^{c}}$ means the generated auxiliary box masks from class $c$, and ${{\tilde{M}}^{c}}$ is the refined CAM of class $c$.

In the final semantic segmentation stage, we adopt dense CRF \cite{krahenbuhl2011efficient} to post-process the refind CAM ${{\tilde{M}}^{c}}$ to get the final pseudo mask ${{M}_{pse}}$, which is used to train a DeepLab V2 \cite{chen2017deeplab} model. The segmentation loss is denoted as ${L}_{seg}$.

\section{Experiments}
\subsection{Experimental Settings}
\subsubsection{Datasets and Evaluation Metric}
The proposed method is evaluated on four egocentric image segmentation benchmarks: EgoHand \cite{bambach2015lending}, EgoHOS-Hands \cite{zhang2022fine}, EgoHOS-HOI \cite{zhang2022fine}, and VISOR-HOS \cite{darkhalil2022epic}. \textbf{EgoHand} is an egocentric hand segmentation dataset, which contains 4800 egocentric frames from 48 video clips for people interacting with their partners in various scenarios. There are 15000 pixel-level annotations of hand instances with 4 categories: my left hand, my right hand, your left hand, and your right hand. In this paper, we split the train and validation sets in a 1: 1 ratio based on the role of actors. \textbf{EgoHOS-Hands} is an egocentric left and right-hand segmentation benchmark. There are 20701 hand instances in 11243 egocentric frames in total, which are collected from Ego4D \cite{grauman2022ego4d}, EPIC-KITCHEN \cite{damen2018scaling}, THU-READ \cite{tang2017action}, and the authors' head-mounted device. In addition, the authors sparsely sample 500 frames on YouTube as an out-of-distribution test set. \textbf{EgoHOS-HOI} is a challenging egocentric hand and interacting object segmentation benchmark. Based on EgoHOS-Hands, EgoHOS-HOI additionally introduces 17568 interacting objects. And there are 100+ subjects and 300+ activities in this benchmark. \textbf{VISOR-HOS} \cite{darkhalil2022epic} is another challenging benchmark, which is also derived from the EPIC-KITCHEN \cite{damen2018scaling}. VISOR-HOS has 32857 images in the \textit{train} set and 7747 in the \textit{val} set from 38 hours of kitchen videos with pixel-level mask annotations. In this paper, we follow the \textit{Hand-Contact-Relation} setting as in \cite{darkhalil2022epic}, which aims to segment the left/right hands and objects in contact. 

In this paper, the mean Intersection over Union (mIoU) metric is used as the evaluation metric for all experiments.

\subsubsection{Implementation Details}
We implement our method using PyTorch. For the initialization of the CLIP model and our visual and text encoder, we adopt the ViT-B/16 CLIP weights \cite{radford2021learning}. Due to limitations in the scale of the dataset, we randomly crop the egocentric frames and assign the multi-label classification labels based on the object's presence. The hyperparameters ${{\lambda }_{1}}$, ${{\lambda }_{2}}$, ${{\lambda }_{3}}$, $\delta $ are set to 5.0, 2.5, 0.2, and 0.025, respectively. In the cognition transferring and decoupling stage,  the model is optimized by AdamW and the initial learning rate is 1e-5 with the polynomial learning rate decay strategy with 32 batch size. The number of epochs is 80 for EgoHand and EgoHOS-Hands, for EgoHOS-HOI and VISOR-HOS, the model is initialized with the parameters trained on EgoHOS-Hands and the number of epochs is 40. Images are resized to 640 $\times$ 320 and no multi-scale strategy is utilized during the CAM generation. The generated CAMs are post-processed by dense CRF \cite{krahenbuhl2011efficient} to generate the final pseudo masks. In the semantic segmentation stage, following the prior WSSS setting \cite{lin2023clip,xie2022clims,lee2021anti}, we use ResNet101-based DeepLab V2 \cite{he2016deep,chen2017deeplab} with cross-entropy loss.

\subsection{Comparison with methods from the related task}

\begin{table*}[!t]
\centering
\caption{Quantitative results of the generated pseudo masks. ``*" denotes the model is frozen, and ``CLIP-ES-FT" means we fine-tune the parameter of the CLIP-ES.}
\label{tab:1}
\scalebox{1.0}{
\begin{tabular}{l|l|p{2.0cm}<{\centering}|p{2.0cm}<{\centering}|p{2.0cm}<{\centering}|p{2.0cm}<{\centering}}
\toprule
Method       & Backbone                  & EgoHand & EgoHOS-Hands & EgoHOS-HOI & VISOR-HOS \\ \hline \hline
CLIP-ES \cite{lin2023clip}     & CLIP(ViT-B/16)*          & 22.76   & 46.49        & 24.56   & 37.81   \\ 
MCTFormer V1 \cite{xu2022multi} & DeiT-S                    & 27.99   & 35.98        & 28.14   & 30.10    \\ 
MCTFormer V2 \cite{xu2022multi} & DeiT-S                    & 32.10   & 45.09        & 25.29    & 35.20  \\ 
MCTFormer+ \cite{xu2023mctformer+} & DeiT-S                    & 28.55   & 43.27        & 21.89  & 35.37    \\ 
AFA \cite{ru2022learning}         & MiT                       & 40.41   & 45.64        & 23.50  & 39.15   \\ 
CLIMS \cite{xie2022clims}       & Res-50+CLIP(ViT-B/32)* & 37.12   & 51.48        & 26.25  & 37.92     \\ 
ToCo \cite{ru2023token}        & ViT-B/16                  & 52.02   & 47.60        & 26.73   &  40.20    \\ 
CLIP-ES-FT \cite{lin2023clip}  & CLIP(ViT-B/16)           & 57.56   & 66.65        & 33.97  & 54.43    \\ 
\textbf{CTDN (Ours)}         & \textbf{CLIP(ViT-B/16)}           & \textbf{72.36}   & \textbf{73.44}      & \textbf{40.01}  & \textbf{61.60}  \\ 
\bottomrule
\end{tabular}}
\end{table*}

\begin{table*}[t]
\centering
\caption{The performance of semantic segmentation. ``\textit{test-I}" and ``\textit{test-O}" denote the in-domain and out-of-domain test sets, respectively. ``V2" denotes the DeepLab V2 \cite{chen2017deeplab} model, \dag denotes the corresponding method uses the end-to-end weakly-supervised semantic segmentation network.}
\label{tab:2}
\scalebox{1.0}{
\begin{tabular}{l|l|p{1.3cm}<{\centering}|ccc|ccc|p{1.5cm}<{\centering}}
\toprule
\multirow{2}{*}{Method} & \multirow{2}{*}{Network} & EgoHand & \multicolumn{3}{c|}{EgoHOS-Hands}                                                & \multicolumn{3}{c|}{EgoHOS-HOI}      & VISOR-HOS                                       \\ \cline{3-10} 
                        &                          & \textit{val}     & \multicolumn{1}{p{1.2cm}<{\centering}|}{\textit{val}}   & \multicolumn{1}{p{1.2cm}<{\centering}|}{\textit{test-I}} & \multicolumn{1}{p{1.2cm}<{\centering}|}{\textit{test-O}} & \multicolumn{1}{p{1.2cm}<{\centering}|}{\textit{val}}   & \multicolumn{1}{p{1.2cm}<{\centering}|}{\textit{test-I}} & \multicolumn{1}{p{1.1cm}<{\centering}|}{\textit{test-O}} & \textit{val} \\ \hline \hline
Fully-supervised                    & Res-101+V2        & 73.80   & \multicolumn{1}{c|}{88.32} & \multicolumn{1}{c|}{88.48}         & 90.18          & \multicolumn{1}{c|}{57.26} & \multicolumn{1}{c|}{58.61}         & 60.94     & 75.47     \\  \hline \hline
CLIP-ES \cite{lin2023clip}                & Res-101+V2        & 22.58   & \multicolumn{1}{c|}{49.15} & \multicolumn{1}{c|}{49.09}         & 51.82          & \multicolumn{1}{c|}{23.49} & \multicolumn{1}{c|}{23.57}         & 23.52   & 37.91       \\ 
MCTFormer V1 \cite{xu2022multi}           & Res-101+V2        & 25.73   & \multicolumn{1}{c|}{32.17} & \multicolumn{1}{c|}{32.00}         & 31.46          & \multicolumn{1}{c|}{25.97} & \multicolumn{1}{c|}{27.25}         & 28.67   & 34.93       \\ 
MCTFormer V2 \cite{xu2022multi}         & Res-101+V2        & 32.65   & \multicolumn{1}{c|}{44.50} & \multicolumn{1}{c|}{44.21}         & 45.45          & \multicolumn{1}{c|}{22.12} & \multicolumn{1}{c|}{22.57}         & 23.69    & 39.41      \\ 
MCTFormer+ \cite{xu2023mctformer+}          & Res-101+V2        & 27.01   & \multicolumn{1}{c|}{42.84} & \multicolumn{1}{c|}{43.24}         & 43.75          & \multicolumn{1}{c|}{20.61} & \multicolumn{1}{c|}{21.18}         & 21.74   & 39.70         \\ 
AFA  \cite{ru2022learning}                   & MiT \dag                    & 34.95   & \multicolumn{1}{c|}{50.29} & \multicolumn{1}{c|}{49.55}              &    51.08           & \multicolumn{1}{c|}{22.37} & \multicolumn{1}{c|}{22.22}              &     22.36     & 41.54    \\ 
CLIMS  \cite{xie2022clims}                & Res-101+V2        & 36.65   & \multicolumn{1}{c|}{50.51} & \multicolumn{1}{c|}{50.27}         & 50.00          & \multicolumn{1}{c|}{24.54} & \multicolumn{1}{c|}{24.40}         & 24.31  & 40.25        \\ 
ToCo \cite{ru2023token}    & ViT-B/16 \dag                & 36.66   & \multicolumn{1}{c|}{45.03} & \multicolumn{1}{c|}{45.32}              &        47.36        & \multicolumn{1}{c|}{23.45} & \multicolumn{1}{c|}{23.86}              &     26.60     & 38.29      \\ 
CLIP-ES-FT \cite{lin2023clip}             & Res-101+V2        & 57.55   & \multicolumn{1}{c|}{74.31} & \multicolumn{1}{c|}{74.56}         & 77.19          & \multicolumn{1}{c|}{37.55} & \multicolumn{1}{c|}{37.56}         & 40.76  & 65.57        \\ 
\textbf{CTDN (Ours)}                  & \textbf{Res-101+V2}        & \textbf{64.45}   & \multicolumn{1}{c|}{\textbf{79.96}} & \multicolumn{1}{c|}{\textbf{80.11}}         & \textbf{85.20}          & \multicolumn{1}{c|}{\textbf{43.34}} & \multicolumn{1}{c|}{\textbf{43.86}}         & \textbf{46.38}   & \textbf{69.90}       \\ 
\bottomrule
\end{tabular}}
\end{table*}

We compare our CTDN with the related third-view weakly-supervised semantic segmentation (WSSS) methods proposed in recent years, i.e. MCTFormer V1 \cite{xu2022multi}, MCTFormer V2 \cite{xu2022multi}, MCTFormer + \cite{xu2023mctformer+}, AFA \cite{ru2022learning},  CLIMS \cite{xie2022clims}, ToCo \cite{ru2023token}, CLIP-ES \cite{lin2023clip}, and the finetuned CLIP-ES \cite{lin2023clip}. The corresponding models are re-implemented using the released official codes on the egocentric benchmarks and apply the same data pre-processing and parameter initialization strategies as in our method to ensure fair comparisons. 

\subsubsection{Quality of generated pseudo masks}
We quantitatively evaluate our CTDN and compare it with the recently released powerful methods from the related WSSS task. The results in Table \ref{tab:1} demonstrate that our method outperforms other WSSS methods by a large margin in the egocentric scenes. Comparing with the methods only based on the single visual modality (i.e. MCTFormer V1 \cite{xu2022multi}, MCTFormer V2 \cite{xu2022multi}, MCTFormer+ \cite{xu2023mctformer+}, AFA \cite{ru2022learning}, and ToCo \cite{ru2023token}), our CTDN achieves substantially higher performance. In particular, ToCo \cite{ru2023token} utilizes patch contrastive learning to alleviate the over-smoothing problem to discriminate the uncertain areas, whereas our method still surpasses it by 20.34$\%$, 25.84$\%$, 13.28$\%$, and 21.40 $\%$ mIoU respectively due to the introduction of the textual cognitive knowledge. CLIMS \cite{xie2022clims} and CLIP-ES \cite{lin2023clip} introduce the text modality and adopt the large-scale vision-language model. CLIP-ES is fully based on the frozen CLIP and the performance is inferior when it comes to egocentric benchmarks because of the ``relation insensitive" and the large distribution gap between different-view images. CLIMS uses the frozen CLIP model to guide a trainable convolutional network to generate CAM, while our method yields higher performance by over 10$\%$ mIoU on four egocentric benchmarks. Finally, we finetune the CLIP-ES model to achieve competitive performance, while our method transfers the cognition and conducts fore/background representation decoupling to mitigate the false activation caused by the interferential objects and achieve higher performance by 14.80$\%$, 6.79$\%$, 6.04$\%$, and 7.17$\%$ in mIoU respectively.

\subsubsection{Segmentation performance}
We perform egocentric semantic segmentation to further evaluate the quality of the generated pseudo masks and the corresponding results are shown in Table \ref{tab:2}. AFA  \cite{ru2022learning} and ToCo \cite{ru2023token} are end-to-end methods, which means the frameworks generate the pseudo masks and perform segmentation simultaneously. Thus we report the segmentation performance of their original frameworks. For other methods, following the common WSSS setting, we use the generated pseudo masks to fully train a ResNet-101-based DeepLab V2 segmentation network \cite{he2016deep,chen2017deeplab} with a cross-entropy loss. In addition, we also perform the segmentation using the ground-truth masks (i.e. Fully-supervised semantic segmentation), which can be considered as the upper bound of TESS. The experimental results demonstrate that our method achieves 64.45$\%$ mIoU on the \textit{val} set of EgoHand, and 79.96$\%$, 80.11$\%$, 85.20$\%$ mIoU on the \textit{val}/\textit{test-I}/\textit{test-O} sets of EgoHOS-Hands, respectively. For the \textit{val}/\textit{test-I}/\textit{test-O} sets of EgoHOS-HOI, our method achieves 43.34$\%$, 43.86$\%$, and 46.38$\%$ mIoU, respectively. Moreover, our method yields 69.90$\%$ mIoU on the \textit{val} set of VISOR-HOS. Our method outperforms other methods for the related task by a large margin, and the performance gap to the fully-supervised semantic segmentation is tolerable. The above results demonstrate that our CTDN is capable of generating high-quality pseudo masks in egocentric scenes, which facilitate the learning of the DeepLab model for accurate semantic segmentation.

\begin{table*}[!t]
\centering
\caption{Ablation results for the key components on the four TESS benchmarks. The mIoU metrics of the generated pseudo masks are presented.}
\label{tab:3}
\begin{tabular}{p{1.0cm}<{\centering}|p{1.0cm}<{\centering}|p{1.0cm}<{\centering}|p{1.0cm}<{\centering}|p{2.0cm}<{\centering}|p{2.0cm}<{\centering}|p{2.0cm}<{\centering}|p{2.0cm}<{\centering}}
\toprule
${p}$ & ${L}_{VRD}$ & ${L}_{CT}$ & ${L}_{FBD}$ & EgoHand & EgoHOS-Hands & EgoHOS-HOI  & VISOR-HOS\\ \hline \hline
\checkmark       & \checkmark    & \checkmark   & \checkmark   &  \textbf{72.36}  & \textbf{73.44}        & \textbf{40.01} & \textbf{61.60}    \\ 
\checkmark     & \checkmark    & \checkmark   &    & 67.64   &       72.23       &     38.03   & 59.88    \\ 
\checkmark     & \checkmark    &    & \checkmark   & 65.34   &     68.27         &     36.03   & 59.59     \\ 
\checkmark     & \checkmark    &    &    & 59.54   &   66.92           &   34.32    & 57.38     \\ 
\checkmark     &     &    &    & 56.67   &      65.76        &    33.36   & 56.78     \\ 
     &     &    &    & 53.69   &     64.26         &   30.95    & 55.99     \\ \bottomrule
\end{tabular}
\end{table*}


\subsubsection{Evaluation on third-view datasets}

\begin{table}[t]
\centering
\caption{Comparison of results of the generated pseudo masks and semantic segmentation on the general third-view Pascal VOC \cite{everingham2010pascal} and MS COCO \cite{lin2014microsoft} datasets.}
\label{tab:5}
\scalebox{0.95}{
\begin{tabular}{p{2.0cm}|p{0.7cm}<{\centering}|p{0.7cm}<{\centering}|p{1.0cm}<{\centering}|p{1.0cm}<{\centering}|p{1.1cm}<{\centering}}
\toprule
\multirow{2}{*}{Method} & \multicolumn{2}{c|}{Pseudo Mask}       & \multicolumn{3}{c}{Semantic Segmentation}      \\ \cline{2-6} 
                        & VOC & COCO & VOC\textit{(val)} & VOC\textit{(test)} & COCO\textit{(val)} \\ \hline \hline
                        SEAM \cite{wang2020self} & 63.6   &  31.5   & 64.5  & 65.7 &  31.9   \\ \hline
                        AFA  \cite{ru2022learning}&  68.7   &  -    & 66.0   & 66.3 &   38.9  \\ \hline                
                        MCTFormer \cite{xu2022multi}&    69.1 &  41.6   &  71.9  & 71.6 & 42.0    \\ \hline
                        CLIMS \cite{xie2022clims} & 70.5    &  -   & 69.3   & 68.7 & -   \\ \hline
                        ToCo \cite{ru2023token} &  73.6   &   -  &  71.1  & 72.2 &  42.3   \\ \hline
                        OAM \cite{chen2023adversarial} &  70.0   &  -  &  70.3  &  69.8  &   -  \\ \hline
                        Qin el al. \cite{qin2024enhanced}& -   & -  &   67.3  &  67.5  &  -   \\ \hline
                        CMPC \cite{wang2024class} &  73.6  & -  &  71.2   &  72.1  &  \textbf{45.9}   \\ \hline
                       \textbf{CTDN (Ours)} &  \textbf{74.2}  &   \textbf{45.0}  &  \textbf{73.9}  & \textbf{74.9}  &   43.9  \\  
                        \bottomrule
\end{tabular}}
\end{table}

To more comprehensively evaluate the effectiveness of the proposed method, we also conduct quantitative experiments on the general third-view Pascal VOC \cite{everingham2010pascal} and MS COCO \cite{lin2014microsoft} datasets, and the mIoU results of pseudo masks and semantic segmentation are shown in Table \ref{tab:5}. Specifically, we re-design the class texts and cognition set in our method to adapt to the third-view scenes and re-train the proposed model to generate the pseudo masks to perform semantic segmentation. The experimental results demonstrate that our method achieves competitive performance and outperforms many recent state-of-the-art general third-view methods. In detail, for the generated pseudo masks, our CTDN outperforms the recent CMPC \cite{wang2024class} by 0.6$\%$ mIoU on the VOC dataset and achieves 45.0$\%$ mIoU on the MS COCO. For the semantic segmentation, the CTDN outperforms the CMPC \cite{wang2024class} by 2.7$\%$ and 2.8$\%$ mIoU on the VOC \textit{val} and \textit{test} sets, respectively. The above experimental results further demonstrate the validity of the proposed method.

\subsection{Ablation Study}
\subsubsection{Analysis of key components}

In Table \ref{tab:3}, we conduct ablation experiments for the CTDN's key components and report the mIoU metrics of the generated pseudo masks. In the experiments, we investigate the effects of the prompt template ${p}$, and ${L}_{VRD}$, ${L}_{CT}$, ${L}_{FBD}$ losses. The first row in Table \ref{tab:3} shows the full CTDN framework performance, which achieves 72.36$\%$, 73.44$\%$, 40.01$\%$, and 61.60$\%$ mIoU on the EgoHand, EgoHOS-Hands, EgoHOS-HOI, and VISOR-HOS benchmarks, respectively. In the second row, we remove the Foreground-background Decoupling Loss ${L}_{FBD}$, and the experimental results show it leads to a 4.72$\%$ mIoU drop on EgoHand. For EgoHOS-Hands and EgoHOS-HOI benchmarks, it causes 1.21$\%$ and 1.98$\%$ mIoU drop. In addition, the ablation of ${L}_{FBD}$ also leads to a performance degradation of 1.72$\%$ mIoU on VISOR-HOS, which demonstrates the effect of the FDM in decoupling the foreground and background representations to mitigate the false activation areas.

Then, we evaluate the effect of ${L}_{CT}$ in the third row in Table \ref{tab:3}. The experimental results show that the performance on all benchmarks is significantly decreased. Specifically, the mIoU metrics drop by 7.02$\%$, 5.17$\%$, 3.98$\%$, and 2.01$\%$ on the four benchmarks, respectively. It demonstrates that it is essential to transfer semantic-level cognitive knowledge into our framework by the proposed CTM for recognizing the foreground and background objects and further decoupling them. In the fourth row, we remove both ${L}_{CT}$ and ${L}_{FBD}$ resulting in a performance decline of 12.82$\%$, 6.52$\%$, 5.69$\%$, and 4.22$\%$ in mIoU compared to the full CTDN respectively, which further verifies the effectiveness of the composition of  ${L}_{FBD}$ and ${L}_{CT}$ in the proposed method.

Next, we evaluate the effect of ${L}_{VRD}$ in the fifth row, which leads to 2.87$\%$, 1.16$\%$, and 0.96$\%$, and 0.60$\%$ mIoU decline, which demonstrates the visual feature-level distillation helps maintain the representational stability and facilitate the following foreground-background disentangling. Finally, we remove the designed prompt template $p$, which leads to 2.98$\%$, 1.50$\%$, 2.41$\%$, and 0.79$\%$ deterioration on four TESS benchmarks. 

\subsubsection{Analysis of hyperparameters}

\begin{figure*}[t]
  \centering
  \begin{subfigure}{0.32\linewidth}
  \centering
  \includegraphics[width=0.95\linewidth]{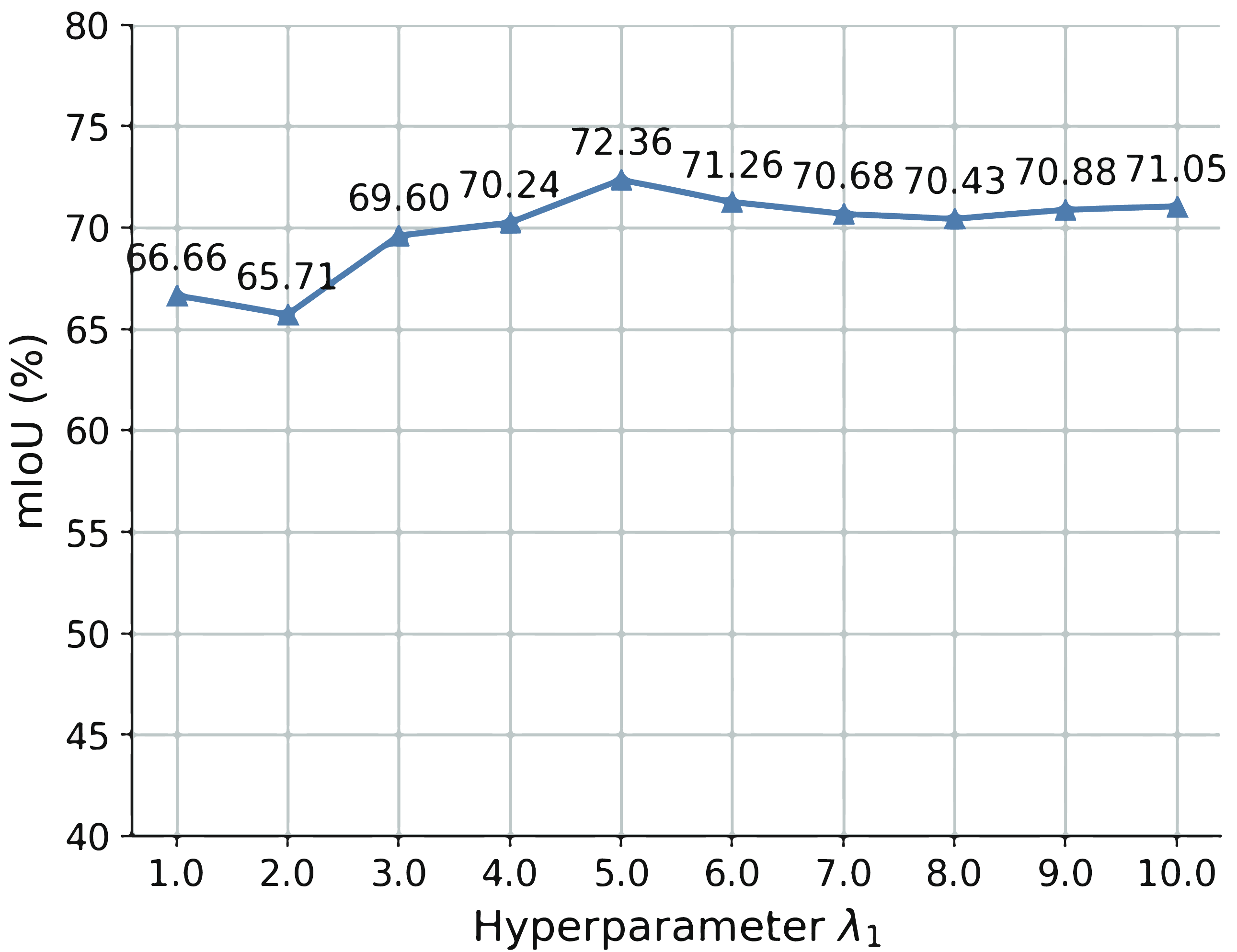}
    \caption{Analysis of ${\lambda}_{1}$}
    \label{fig:4-a}
  \end{subfigure}
  \begin{subfigure}{0.32\linewidth}
  \centering
  \includegraphics[width=0.95\linewidth]{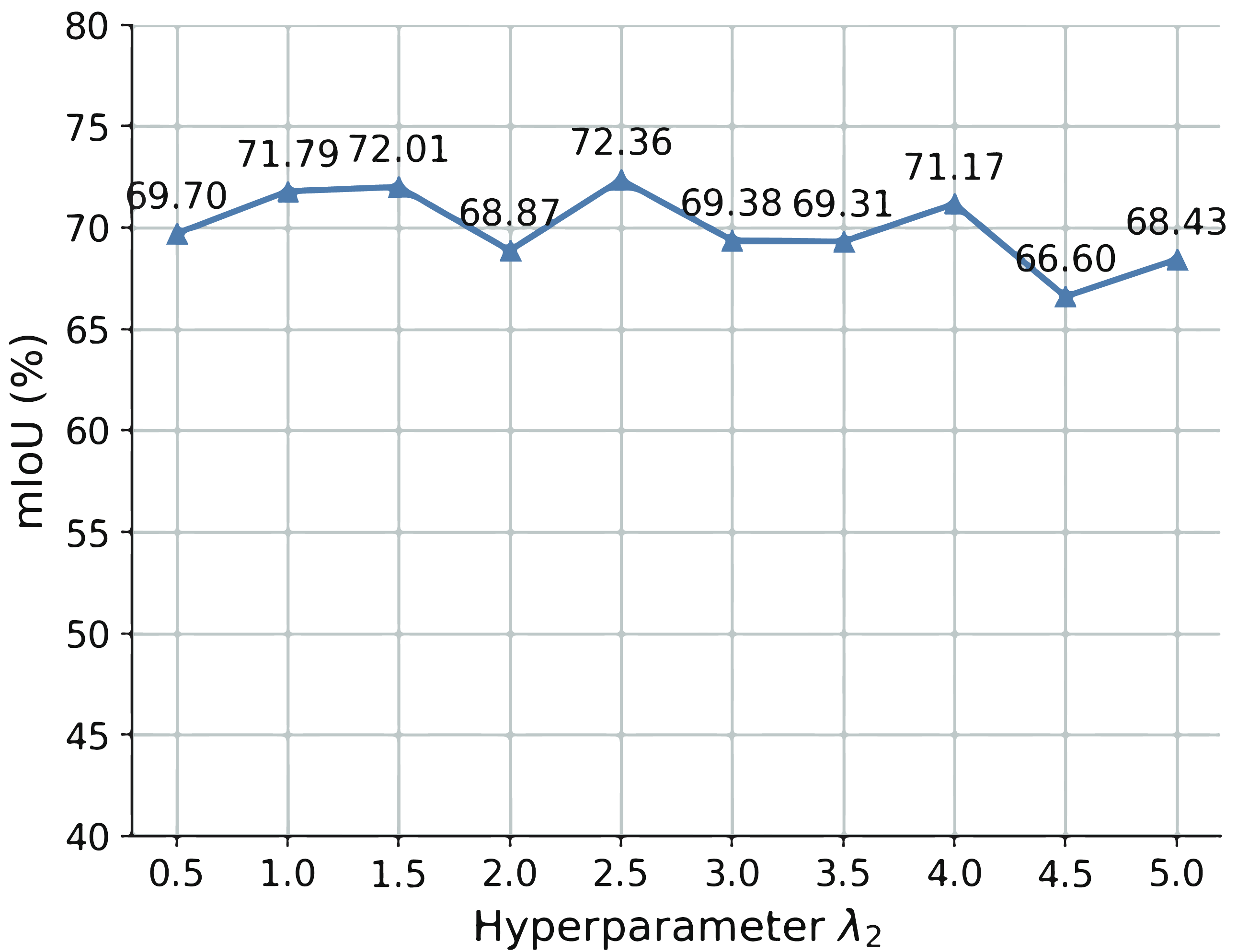}
    \caption{Analysis of ${\lambda}_{2}$}
    \label{fig:4-b}
  \end{subfigure}
  \begin{subfigure}{0.32\linewidth}
  \centering
  \includegraphics[width=0.95\linewidth]{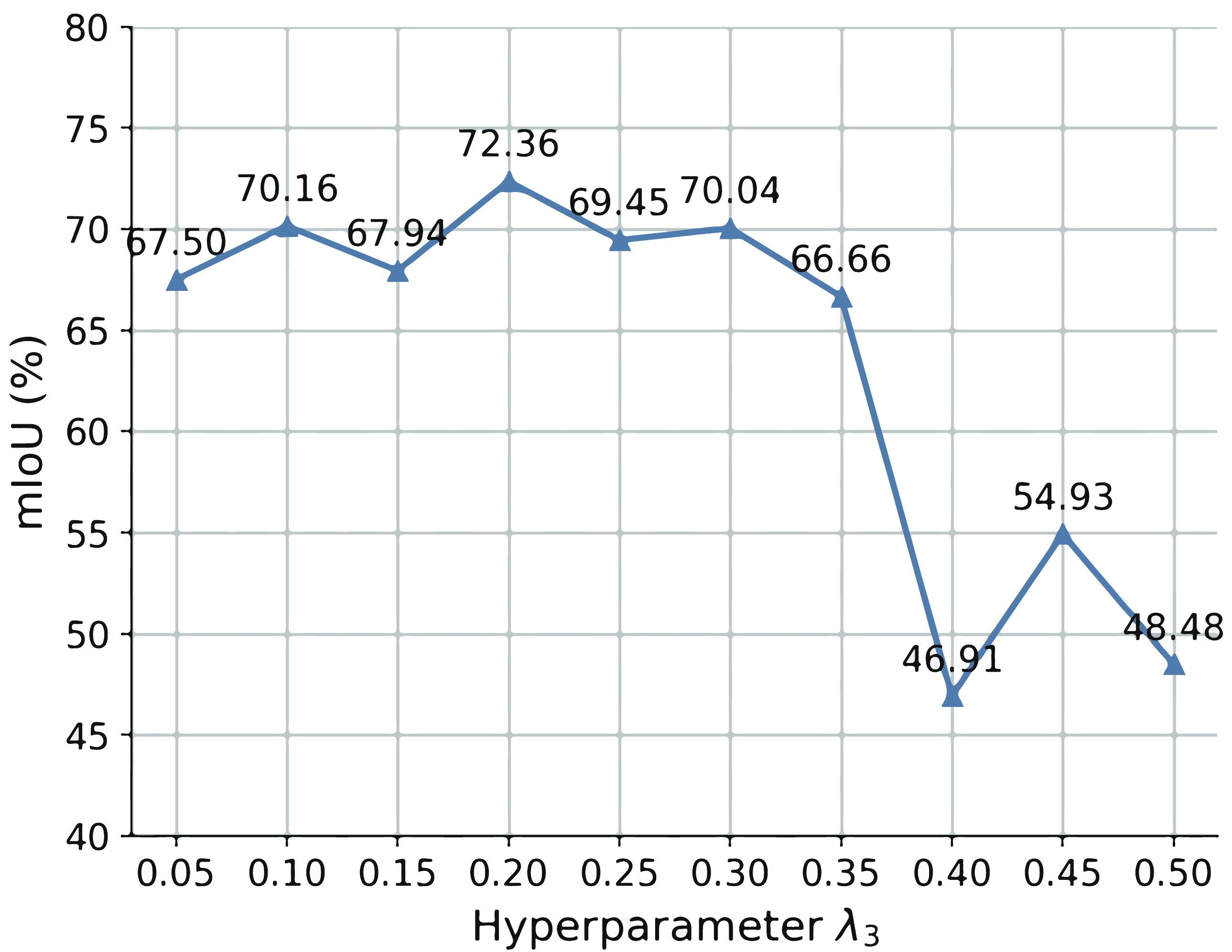}
    \caption{Analysis of ${\lambda}_{3}$}
    \label{fig:4-c}
  \end{subfigure}
  \hfill
  \caption{Analysis on hyperparameters ${\lambda}_{1}$, ${\lambda}_{2}$, and ${\lambda}_{3}$ on the Egohand benchmark.}
  \label{fig:4}
\end{figure*}

\begin{figure}[t]
\centering
\includegraphics[width=1.0\linewidth]{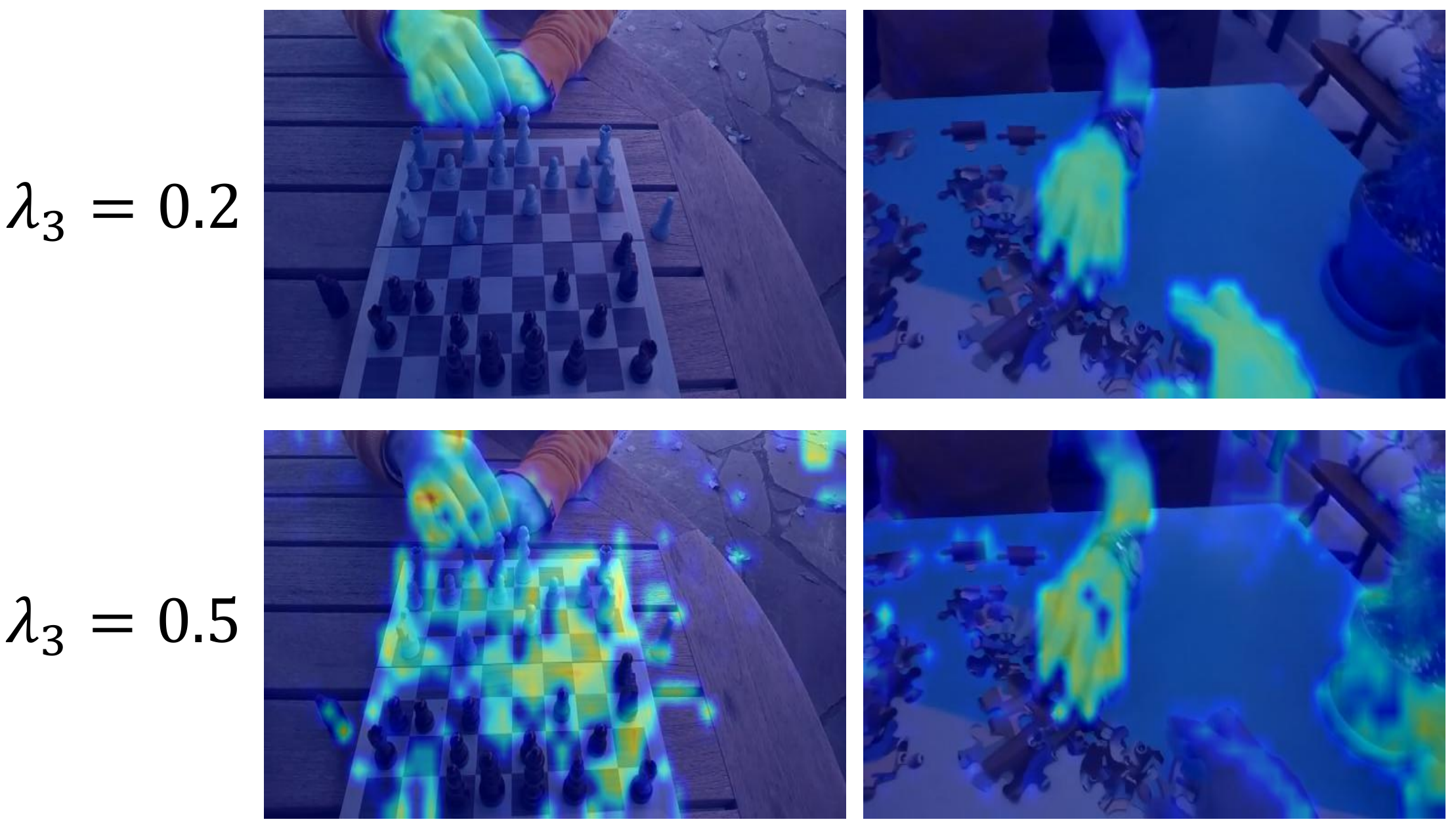}
\caption{The visualizations of the initial class-activation maps (CAM) under different hyperparameter ${\lambda}_{3}$ settings on the EgoHand benchmark.}
\label{fig:8}
\end{figure}

In Fig. \ref{fig:4}, we analyze the hyperparameters ${\lambda}_{1}$, ${\lambda}_{2}$, and ${\lambda}_{3}$ and report the mIoU of the generated pseudo masks. In Fig. \ref{fig:4-a}, when the value of ${\lambda}_{1}$ increases from 1.0 to 10.0, the performance fluctuates within 65.71$\%$ and 72.36$\%$. And when ${\lambda}_{1}$ is set to 5.0, our method achieves the highest performance. Fig. \ref{fig:4-b} shows the mIoU metrics vary within the range of 5.76$\%$ when the ${\lambda}_{2}$ grows from 0.5 to 5.0. The performance peaks when the ${\lambda}_{2}$ is set to 2.5. In Fig. \ref{fig:4-c}, when the value of ${\lambda}_{3}$ increases from 0.05 to 0.35, the mIoU fluctuates within a relatively small range. When the ${\lambda}_{3}$ increases above 0.35, the performance drops remarkably. And the value of  ${\lambda}_{3}$ is preferably set to 0.2.

The performance rapidly declines as the parameter ${\lambda}_{3}$ increases because a high ${\lambda}_{3}$ may causes the model to over-focus on the contrastive representation learning, thereby further affecting the essential egocentric relation learning and cognitive knowledge transferring. In detail, when ${\lambda}_{3}$ is overly high, the value of foreground-background decoupling loss ${L}_{FBD}$ is remarkably higher than those of the relation learning loss ${{L}_{rel}}$ and cognition transferring losses ${L}_{CT}$, ${L}_{VRD}$. In detail, ${{L}_{rel}}$ correlates egocentric images and class texts with dense relations, ${L}_{CT}$ and ${L}_{VRD}$ conduct knowledge distillation of visual features and cognition-based cross-modal logits, respectively. It is easy to raise difficulties for the model in establishing the cross-modal consistency between the egocentric visual objects and class or cognition texts, which are crucial for accurately generating CAM and pseudo masks. As shown in Fig. \ref{fig:8}, compared to the CAM with ${\lambda}_{3}=0.2$, when ${\lambda}_{3}=0.5$, the lack of the visual-textual cross-modal consistency leads to many false positive or false negative regions in the generated CAM.

\subsubsection{Analysis of trainable or frozen text encoder}

\begin{table*}[t]
\centering
\caption{The performance of the generated pseudo masks of our CTDN with trainable or frozen text encoder.}
\label{tab:4}
\scalebox{1.0}{
\begin{tabular}{l|p{2.0cm}<{\centering}|p{2.0cm}<{\centering}|p{2.0cm}<{\centering}|p{2.0cm}<{\centering}}
\toprule
Text Encoder   & EgoHand & EgoHOS-Hands & EgoHOS-HOI & VISOR-HOS \\ \hline \hline
Trainable             &  64.87  &   66.87    &  34.89  &  58.97 \\ \hline
Frozen            & \textbf{72.36}   & \textbf{73.44}      & \textbf{40.01}  & \textbf{61.60}  \\ 
\bottomrule
\end{tabular}}
\end{table*}

We conduct experiments to analyze the effectiveness of the frozen or trainable text encoder in the first Cognition Transferring and Decoupling stage, and the performance of the generated pseudo masks is presented in Table \ref{tab:4}. The experimental results show that the model using the frozen text encoder in the training process achieves better performance. In detail, compared to the trainable text encoder, the performance of the generated pseudo mask corresponding to the frozen text encoder increases by 7.49$\%$, 6.57$\%$, 5.12$\%$, and 2.63$\%$ mIoU on the four TESS benchmarks, respectively. The above results demonstrate that compared to the trainable text encoder, the frozen text encoder is more effective in accurately extracting features of texts in the class set, foreground set, and background set, which facilitates the learning of the dense egocentric relations and egocentric visual representations for generating high-quality pseudo masks.

\subsection{In-depth Analysis}
\subsubsection{Analysis of pseudo masks}

\begin{figure*}[!t]
\centering
\includegraphics[width=1.0\linewidth]{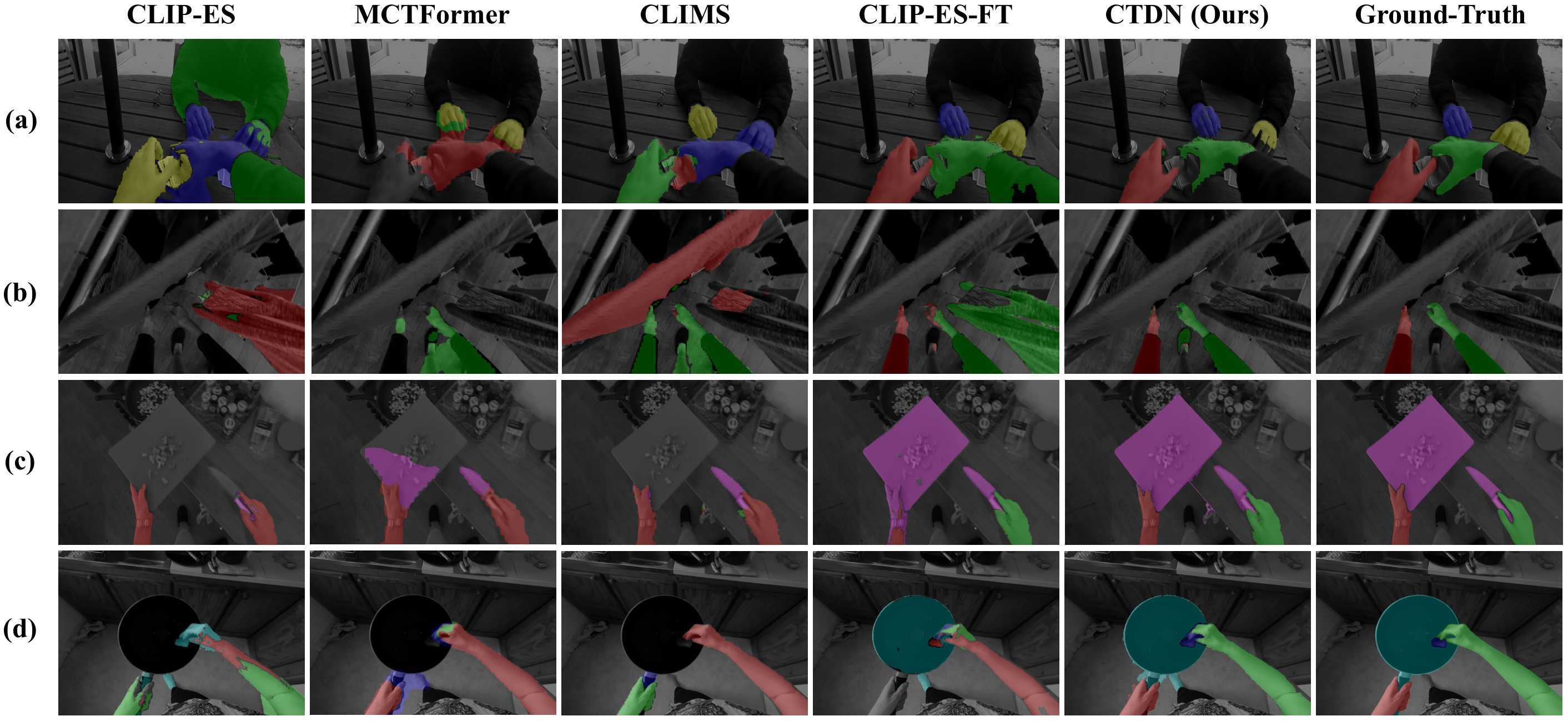}
\caption{Visualization examples of the ground-truths and pseudo masks generated by comparison methods and our CTDN.}
\label{fig:5}
\end{figure*}

\begin{figure*}[t]
\centering
\includegraphics[width=1.0\linewidth]{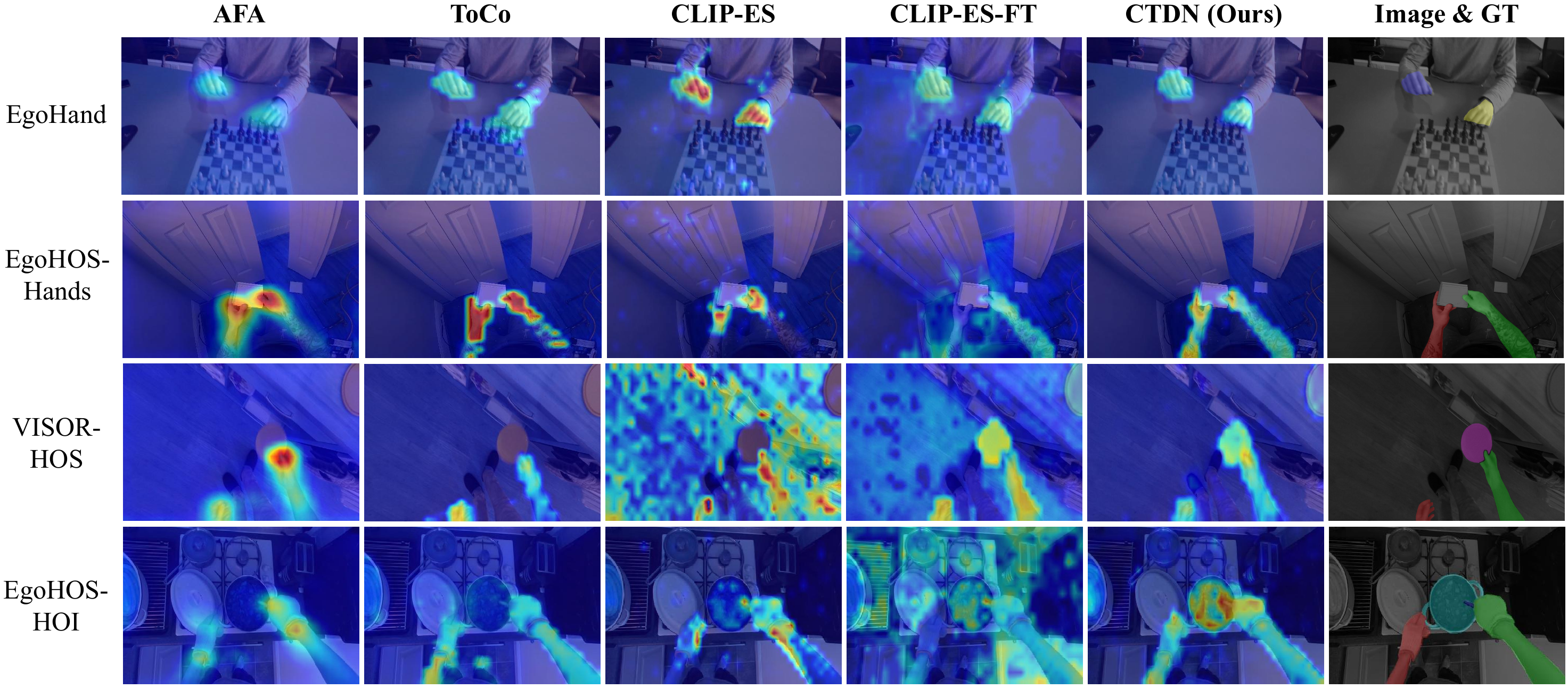}
\caption{Comparison of the initial class-activation map (CAM) generated by CTDN and other comparison methods.}
\label{fig:6}
\end{figure*}

\begin{figure*}[t]
\centering
\includegraphics[width=1.0\linewidth]{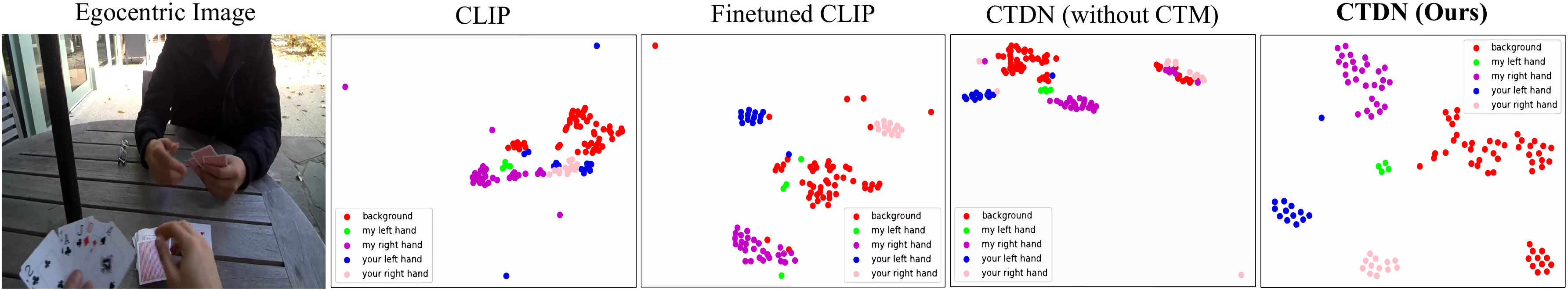}
\caption{Visualizations of the visual representations extracted by different models. (Best viewed in color).}
\label{fig:7}
\end{figure*}

In Fig. \ref{fig:5}, we present several examples of the ground-truth segmentation masks and the generated pseudo masks of different methods on the four TESS benchmarks for qualitative evaluation.

Row (a) shows the example of the EgoHand benchmark, which aims to segment the hands of the egocentric camera wearer and partner. CLIP-ES \cite{lin2023clip} incorrectly segment the partner's body, and MCTFormer \cite{xu2022multi} and CLIMS \cite{xie2022clims} miscategorize the segmented hands. The large false positive areas exist in the result of the finetuned CLIP-ES-FT \cite{xu2022multi}. While our CTDN accurately segments and categorizes all hands in the egocentric image without false positive regions. Row (b) is an example of EgoHOS-Hands, which aims to segment the left and right hands and arms. Other comparison methods either miscategorize the left and right hands and arms or generate pseudo masks with severe noise. In contrast, our method is capable of identifying the side of the hands and generates an accurate pseudo mask with tolerable noises, which is better than the results of other methods. The example of the VISOR-HOS benchmark, which aims to segment the left/right hands and the object in contact, is shown in row (c). CLIP-ES miscategorizes the hands and fails in segmenting the cutting board and knife. MCTFormer and CLIMS only segment a portion of areas of the contacting objects. CLIP-ES-FT confuses the left hand and the in-contact cutting board, resulting in a large region of the left hand miscategorized. Row (d) shows an example of the EgoHOS-HOI benchmark, which is more challenging than the VISOR-HOS benchmark because it requires distinguishing the objects in the left/right/both hand(s). Compared with other comparison methods, our CTDN generates a complete and accurate pseudo mask and can precisely segment the tiny object (i.e. sponge in right hand).

\subsubsection{Analysis of CAM}
The initial class-activation map (CAM) is essential to the quality of the final generated pseudo masks. Therefore, we visualize the initial CAM generated by our CTDN and other comparison methods \cite{ru2022learning,ru2023token,lin2023clip} on the four TESS benchmarks. 

As shown in Fig. \ref{fig:6}, AFA \cite{ru2022learning} and ToCo \cite{ru2023token} roughly localize the hands for simple egocentric hand segmentation with fuzzy boundaries. Under the more challenging VISOR-HOS and EgoHOS-HOI in the third and fourth rows, they fail to highlight the plate and pan in contact with the hands. CLIP-ES \cite{lin2023clip} incompletely highlights a part region of the hands and fails to highlight the interacting object in the fourth row. Moreover, CLIP-ES produces severe noises in the third row, which impairs the quality of generated pseudo masks. The finetuned CLIP-ES-FT framework is confused by the foreground-background interferential objects and incorrectly highlights the regions of background objects such as the desk and gas stove. In the fifth column, the proposed CTDN generates complete and accurate CAM with clear boundaries and tolerable noises, which demonstrates the effect of the transferred cognitive knowledge and the foreground-background representation decoupling.

\subsubsection{Analysis of foreground-background decoupling}
The proposed method transfers the cognitive knowledge into the egocentric framework, which facilitates the representation decoupling of the foreground and background regions. To intuitively present the disentangled foreground/background visual representations, as shown in Fig. \ref{fig:7}, we use the t-SNE \cite{van2008visualizing} to project the visual representations into 2-D planes. We use colored dots to indicate representations of different categories. 

The first column shows the egocentric image where the camera wearer and partner are playing poker, and the model is required to learn visual representations and generate the pseudo mask for distinct hand regions. As shown in the second column in Fig. \ref{fig:7}, the visual representations of different categories extracted by the original CLIP model aggregate with unclear decision boundaries, which causes many miscategorized regions in the generated pseudo masks. The third column shows that although the finetuned CLIP model roughly discriminates different classes, the distance between the foreground and background representations is close, and many background representations anomalously exist in the foreground clusters (i.e. foreground-background interference). The above situation is also encountered in the fourth column because the model cannot recognize fore/background objects without the transferred cognitive knowledge. Finally, in the fifth column, the representations of different categories are explicitly disentangled with clear inter-category decision boundaries, and representations within the same category are tightly aggregated into clusters, which demonstrates the effectiveness of our method.

\section{Limitations and Future Works}
Our method has addressed dense wearer-object relations and complex inter-object interference in egocentric scenes. Specifically, we correlate egocentric images and class texts and transfer cognitive knowledge to disentangle the foreground and background representations, leading to remarkable performance on multiple TESS benchmarks. In addition, the challenge of accurately segmenting objects that are motion-blurred or undergoing state changes remains in egocentric contexts. In the future, we will explore some approaches for temporal modeling in egocentric scenes to overcome the challenge. On the one hand, we will apply video backbones as the visual encoder to accommodate egocentric video clips as input. On the other hand, we will incorporate temporal attention mechanisms or optical flow-based approaches into our framework for inter-frame correlation modeling of egocentric videos.

\section{Conclusion}
In this paper, we explore a novel Text-supervised Egocentric Semantic Segmentation (TESS) task that aims to utilize weak image-level text supervision to perform egocentric semantic segmentation considering the cumbersome pixel-level labeling problem. To support the research, we propose a Cognition Transferring and Decoupling Network (CTDN), which learns the egocentric wearer-object relations and transfers the cognitive knowledge from the large-scale pre-trained model into our egocentric model via a Cognition Transferring Module (CTM). Then, the Foreground-background Decoupling Module (FDM) is adopted to disentangle the visual representations based on the transferred cognition for explicitly discriminating the foreground and background regions to mitigate false activation areas. Extensive experiments demonstrate the effectiveness of our method and its superiority over many related methods.

\bibliographystyle{IEEEtran}
\bibliography{references}{}

\begin{thebibliography}{100}
\providecommand{\url}[1]{#1}
\csname url@samestyle\endcsname
\providecommand{\newblock}{\relax}
\providecommand{\bibinfo}[2]{#2}
\providecommand{\BIBentrySTDinterwordspacing}{\spaceskip=0pt\relax}
\providecommand{\BIBentryALTinterwordstretchfactor}{4}
\providecommand{\BIBentryALTinterwordspacing}{\spaceskip=\fontdimen2\font plus
\BIBentryALTinterwordstretchfactor\fontdimen3\font minus \fontdimen4\font\relax}
\providecommand{\BIBforeignlanguage}[2]{{%
\expandafter\ifx\csname l@#1\endcsname\relax
\typeout{** WARNING: IEEEtran.bst: No hyphenation pattern has been}%
\typeout{** loaded for the language `#1'. Using the pattern for}%
\typeout{** the default language instead.}%
\else
\language=\csname l@#1\endcsname
\fi
#2}}
\providecommand{\BIBdecl}{\relax}
\BIBdecl

\bibitem{bambach2015lending}
S.~Bambach, S.~Lee, D.~J. Crandall, and C.~Yu, ``Lending a hand: Detecting hands and recognizing activities in complex egocentric interactions,'' in \emph{Proceedings of the IEEE international conference on computer vision}, 2015, pp. 1949--1957.

\bibitem{zhang2022fine}
L.~Zhang, S.~Zhou, S.~Stent, and J.~Shi, ``Fine-grained egocentric hand-object segmentation: Dataset, model, and applications,'' in \emph{European Conference on Computer Vision}.\hskip 1em plus 0.5em minus 0.4em\relax Springer, 2022, pp. 127--145.

\bibitem{jia2022generative}
W.~Jia, M.~Liu, and J.~M. Rehg, ``Generative adversarial network for future hand segmentation from egocentric video,'' in \emph{European Conference on Computer Vision}.\hskip 1em plus 0.5em minus 0.4em\relax Springer, 2022, pp. 639--656.

\bibitem{cai2020generalizing}
M.~Cai, F.~Lu, and Y.~Sato, ``Generalizing hand segmentation in egocentric videos with uncertainty-guided model adaptation,'' in \emph{Proceedings of the ieee/cvf conference on computer vision and pattern recognition}, 2020, pp. 14\,392--14\,401.

\bibitem{ji2019survey}
Y.~Ji, Y.~Yang, F.~Shen, H.~T. Shen, and X.~Li, ``A survey of human action analysis in hri applications,'' \emph{IEEE Transactions on Circuits and Systems for Video Technology}, vol.~30, no.~7, pp. 2114--2128, 2019.

\bibitem{ji2020arbitrary}
Y.~Ji, Y.~Yang, F.~Shen, H.~T. Shen, and W.-S. Zheng, ``Arbitrary-view human action recognition: A varying-view rgb-d action dataset,'' \emph{IEEE Transactions on Circuits and Systems for Video Technology}, vol.~31, no.~1, pp. 289--300, 2020.

\bibitem{li2021survey}
R.~Li, H.~Wang, and Z.~Liu, ``Survey on mapping human hand motion to robotic hands for teleoperation,'' \emph{IEEE Transactions on Circuits and Systems for Video Technology}, vol.~32, no.~5, pp. 2647--2665, 2021.

\bibitem{li2023proactive}
S.~Li, P.~Zheng, S.~Liu, Z.~Wang, X.~V. Wang, L.~Zheng, and L.~Wang, ``Proactive human--robot collaboration: Mutual-cognitive, predictable, and self-organising perspectives,'' \emph{Robotics and Computer-Integrated Manufacturing}, vol.~81, p. 102510, 2023.

\bibitem{dahiya2023survey}
A.~Dahiya, A.~M. Aroyo, K.~Dautenhahn, and S.~L. Smith, ``A survey of multi-agent human--robot interaction systems,'' \emph{Robotics and Autonomous Systems}, vol. 161, p. 104335, 2023.

\bibitem{Qiu_2024_CVPR}
H.~Qiu, L.~Wang, T.~Zhao, F.~Meng, and H.~Li, ``Humanformer: Human-centric prompting multi-modal perception transformer for referring crowd detection,'' in \emph{Proceedings of the IEEE/CVF Conference on Computer Vision and Pattern Recognition (CVPR) Workshops}, June 2024, pp. 5530--5540.

\bibitem{zhang2020language}
W.~Zhang, C.~Ma, Q.~Wu, and X.~Yang, ``Language-guided navigation via cross-modal grounding and alternate adversarial learning,'' \emph{IEEE Transactions on Circuits and Systems for Video Technology}, vol.~31, no.~9, pp. 3469--3481, 2020.

\bibitem{pu2023rules}
J.~Pu, H.~Duan, J.~Zhao, and Y.~Long, ``Rules for expectation: Learning to generate rules via social environment modelling,'' \emph{IEEE Transactions on Circuits and Systems for Video Technology}, 2023.

\bibitem{nahavandi2022application}
D.~Nahavandi, R.~Alizadehsani, A.~Khosravi, and U.~R. Acharya, ``Application of artificial intelligence in wearable devices: Opportunities and challenges,'' \emph{Computer Methods and Programs in Biomedicine}, vol. 213, p. 106541, 2022.

\bibitem{he2023learning}
Z.~He, L.~Wang, R.~Dang, S.~Li, Q.~Yan, C.~Liu, and Q.~Chen, ``Learning depth representation from rgb-d videos by time-aware contrastive pre-training,'' \emph{IEEE Transactions on Circuits and Systems for Video Technology}, 2023.

\bibitem{zhan2024enhancing}
Z.~Zhan, J.~Qin, W.~Zhuo, and G.~Tan, ``Enhancing vision and language navigation with prompt-based scene knowledge,'' \emph{IEEE Transactions on Circuits and Systems for Video Technology}, 2024.

\bibitem{lin2020ego2hands}
F.~Lin, B.~Price, and T.~Martinez, ``Ego2hands: A dataset for egocentric two-hand segmentation and detection,'' \emph{arXiv preprint arXiv:2011.07252}, 2020.

\bibitem{gonzalez2020enhanced}
E.~Gonzalez-Sosa, P.~Perez, R.~Tolosana, R.~Kachach, and A.~Villegas, ``Enhanced self-perception in mixed reality: Egocentric arm segmentation and database with automatic labeling,'' \emph{IEEE Access}, vol.~8, pp. 146\,887--146\,900, 2020.

\bibitem{yuan2021simple}
J.~Yuan, Y.~Liu, C.~Shen, Z.~Wang, and H.~Li, ``A simple baseline for semi-supervised semantic segmentation with strong data augmentation,'' in \emph{Proceedings of the IEEE/CVF International Conference on Computer Vision}, 2021, pp. 8229--8238.

\bibitem{ren2022adela}
H.~Ren, Y.~Yang, H.~Wang, B.~Shen, Q.~Fan, Y.~Zheng, C.~K. Liu, and L.~J. Guibas, ``Adela: Automatic dense labeling with attention for viewpoint shift in semantic segmentation,'' in \emph{Proceedings of the IEEE/CVF Conference on Computer Vision and Pattern Recognition}, 2022, pp. 8079--8089.

\bibitem{li2019supervised}
Y.~Li, L.~Jia, Z.~Wang, Y.~Qian, and H.~Qiao, ``Un-supervised and semi-supervised hand segmentation in egocentric images with noisy label learning,'' \emph{Neurocomputing}, vol. 334, pp. 11--24, 2019.

\bibitem{wu2023continual}
W.~Wu, Y.~Zhao, Z.~Li, L.~Shan, H.~Zhou, and M.~Z. Shou, ``Continual learning for image segmentation with dynamic query,'' \emph{IEEE Transactions on Circuits and Systems for Video Technology}, 2023.

\bibitem{wei2016stc}
Y.~Wei, X.~Liang, Y.~Chen, X.~Shen, M.-M. Cheng, J.~Feng, Y.~Zhao, and S.~Yan, ``Stc: A simple to complex framework for weakly-supervised semantic segmentation,'' \emph{IEEE transactions on pattern analysis and machine intelligence}, vol.~39, no.~11, pp. 2314--2320, 2016.

\bibitem{kolesnikov2016seed}
A.~Kolesnikov and C.~H. Lampert, ``Seed, expand and constrain: Three principles for weakly-supervised image segmentation,'' in \emph{Computer Vision--ECCV 2016: 14th European Conference, Amsterdam, The Netherlands, October 11--14, 2016, Proceedings, Part IV 14}.\hskip 1em plus 0.5em minus 0.4em\relax Springer, 2016, pp. 695--711.

\bibitem{ahn2018learning}
J.~Ahn and S.~Kwak, ``Learning pixel-level semantic affinity with image-level supervision for weakly supervised semantic segmentation,'' in \emph{Proceedings of the IEEE conference on computer vision and pattern recognition}, 2018, pp. 4981--4990.

\bibitem{xu2022multi}
L.~Xu, W.~Ouyang, M.~Bennamoun, F.~Boussaid, and D.~Xu, ``Multi-class token transformer for weakly supervised semantic segmentation,'' in \emph{Proceedings of the IEEE/CVF Conference on Computer Vision and Pattern Recognition}, 2022, pp. 4310--4319.

\bibitem{Xie_2022_CVPR}
J.~Xie, J.~Xiang, J.~Chen, X.~Hou, X.~Zhao, and L.~Shen, ``C2am: Contrastive learning of class-agnostic activation map for weakly supervised object localization and semantic segmentation,'' in \emph{Proceedings of the IEEE/CVF Conference on Computer Vision and Pattern Recognition (CVPR)}, June 2022, pp. 989--998.

\bibitem{ru2022learning}
L.~Ru, Y.~Zhan, B.~Yu, and B.~Du, ``Learning affinity from attention: End-to-end weakly-supervised semantic segmentation with transformers,'' in \emph{Proceedings of the IEEE/CVF Conference on Computer Vision and Pattern Recognition}, 2022, pp. 16\,846--16\,855.

\bibitem{ru2023token}
L.~Ru, H.~Zheng, Y.~Zhan, and B.~Du, ``Token contrast for weakly-supervised semantic segmentation,'' in \emph{Proceedings of the IEEE/CVF Conference on Computer Vision and Pattern Recognition}, 2023, pp. 3093--3102.

\bibitem{xu2023mctformer+}
L.~Xu, M.~Bennamoun, F.~Boussaid, H.~Laga, W.~Ouyang, and D.~Xu, ``Mctformer+: Multi-class token transformer for weakly supervised semantic segmentation,'' \emph{IEEE transactions on pattern analysis and machine intelligence}, 2024.

\bibitem{xie2022clims}
J.~Xie, X.~Hou, K.~Ye, and L.~Shen, ``Clims: Cross language image matching for weakly supervised semantic segmentation,'' in \emph{Proceedings of the IEEE/CVF Conference on Computer Vision and Pattern Recognition}, 2022, pp. 4483--4492.

\bibitem{radford2021learning}
A.~Radford, J.~W. Kim, C.~Hallacy, A.~Ramesh, G.~Goh, S.~Agarwal, G.~Sastry, A.~Askell, P.~Mishkin, J.~Clark \emph{et~al.}, ``Learning transferable visual models from natural language supervision,'' in \emph{International conference on machine learning}.\hskip 1em plus 0.5em minus 0.4em\relax PMLR, 2021, pp. 8748--8763.

\bibitem{jia2021scaling}
C.~Jia, Y.~Yang, Y.~Xia, Y.-T. Chen, Z.~Parekh, H.~Pham, Q.~Le, Y.-H. Sung, Z.~Li, and T.~Duerig, ``Scaling up visual and vision-language representation learning with noisy text supervision,'' in \emph{International conference on machine learning}.\hskip 1em plus 0.5em minus 0.4em\relax PMLR, 2021, pp. 4904--4916.

\bibitem{nilsson2021embodied}
D.~Nilsson, A.~Pirinen, E.~G{\"a}rtner, and C.~Sminchisescu, ``Embodied visual active learning for semantic segmentation,'' in \emph{Proceedings of the AAAI Conference on Artificial Intelligence}, vol.~35, no.~3, 2021, pp. 2373--2383.

\bibitem{darkhalil2022epic}
A.~Darkhalil, D.~Shan, B.~Zhu, J.~Ma, A.~Kar, R.~Higgins, S.~Fidler, D.~Fouhey, and D.~Damen, ``Epic-kitchens visor benchmark: Video segmentations and object relations,'' \emph{Advances in Neural Information Processing Systems}, vol.~35, pp. 13\,745--13\,758, 2022.

\bibitem{lin2023clip}
Y.~Lin, M.~Chen, W.~Wang, B.~Wu, K.~Li, B.~Lin, H.~Liu, and X.~He, ``Clip is also an efficient segmenter: A text-driven approach for weakly supervised semantic segmentation,'' in \emph{Proceedings of the IEEE/CVF Conference on Computer Vision and Pattern Recognition}, 2023, pp. 15\,305--15\,314.

\bibitem{gu2021open}
X.~Gu, T.-Y. Lin, W.~Kuo, and Y.~Cui, ``Open-vocabulary object detection via vision and language knowledge distillation,'' \emph{arXiv preprint arXiv:2104.13921}, 2021.

\bibitem{wang2022clip}
Z.~Wang, N.~Codella, Y.-C. Chen, L.~Zhou, J.~Yang, X.~Dai, B.~Xiao, H.~You, S.-F. Chang, and L.~Yuan, ``Clip-td: Clip targeted distillation for vision-language tasks,'' \emph{arXiv preprint arXiv:2201.05729}, 2022.

\bibitem{wu2023clipself}
S.~Wu, W.~Zhang, L.~Xu, S.~Jin, X.~Li, W.~Liu, and C.~C. Loy, ``Clipself: Vision transformer distills itself for open-vocabulary dense prediction,'' \emph{arXiv preprint arXiv:2310.01403}, 2023.

\bibitem{chen2023exploring}
J.~Chen, D.~Zhu, G.~Qian, B.~Ghanem, Z.~Yan, C.~Zhu, F.~Xiao, S.~C. Culatana, and M.~Elhoseiny, ``Exploring open-vocabulary semantic segmentation from clip vision encoder distillation only,'' in \emph{Proceedings of the IEEE/CVF International Conference on Computer Vision}, 2023, pp. 699--710.

\bibitem{wu2023tinyclip}
K.~Wu, H.~Peng, Z.~Zhou, B.~Xiao, M.~Liu, L.~Yuan, H.~Xuan, M.~Valenzuela, X.~S. Chen, X.~Wang \emph{et~al.}, ``Tinyclip: Clip distillation via affinity mimicking and weight inheritance,'' in \emph{Proceedings of the IEEE/CVF International Conference on Computer Vision}, 2023, pp. 21\,970--21\,980.

\bibitem{rasheed2023fine}
H.~Rasheed, M.~U. Khattak, M.~Maaz, S.~Khan, and F.~S. Khan, ``Fine-tuned clip models are efficient video learners,'' in \emph{Proceedings of the IEEE/CVF Conference on Computer Vision and Pattern Recognition}, 2023, pp. 6545--6554.

\bibitem{kong2022human}
Y.~Kong and Y.~Fu, ``Human action recognition and prediction: A survey,'' \emph{International Journal of Computer Vision}, vol. 130, no.~5, pp. 1366--1401, 2022.

\bibitem{ji20123d}
S.~Ji, W.~Xu, M.~Yang, and K.~Yu, ``3d convolutional neural networks for human action recognition,'' \emph{IEEE transactions on pattern analysis and machine intelligence}, vol.~35, no.~1, pp. 221--231, 2012.

\bibitem{yao2019review}
G.~Yao, T.~Lei, and J.~Zhong, ``A review of convolutional-neural-network-based action recognition,'' \emph{Pattern Recognition Letters}, vol. 118, pp. 14--22, 2019.

\bibitem{xing2023svformer}
Z.~Xing, Q.~Dai, H.~Hu, J.~Chen, Z.~Wu, and Y.-G. Jiang, ``Svformer: Semi-supervised video transformer for action recognition,'' in \emph{Proceedings of the IEEE/CVF Conference on Computer Vision and Pattern Recognition}, 2023, pp. 18\,816--18\,826.

\bibitem{cao2023vs}
C.~Cao, Z.~Sun, Q.~Lv, L.~Min, and Y.~Zhang, ``Vs-transgru: A novel transformer-gru-based framework enhanced by visual-semantic fusion for egocentric action anticipation,'' \emph{arXiv preprint arXiv:2307.03918}, 2023.

\bibitem{grauman2022ego4d}
K.~Grauman, A.~Westbury, E.~Byrne, Z.~Chavis, A.~Furnari, R.~Girdhar, J.~Hamburger, H.~Jiang, M.~Liu, X.~Liu \emph{et~al.}, ``Ego4d: Around the world in 3,000 hours of egocentric video,'' in \emph{Proceedings of the IEEE/CVF Conference on Computer Vision and Pattern Recognition}, 2022, pp. 18\,995--19\,012.

\bibitem{damen2018scaling}
D.~Damen, H.~Doughty, G.~M. Farinella, S.~Fidler, A.~Furnari, E.~Kazakos, D.~Moltisanti, J.~Munro, T.~Perrett, W.~Price \emph{et~al.}, ``Scaling egocentric vision: The epic-kitchens dataset,'' in \emph{Proceedings of the European conference on computer vision (ECCV)}, 2018, pp. 720--736.

\bibitem{sigurdsson2018charades}
G.~A. Sigurdsson, A.~Gupta, C.~Schmid, A.~Farhadi, and K.~Alahari, ``Charades-ego: A large-scale dataset of paired third and first person videos,'' \emph{arXiv preprint arXiv:1804.09626}, 2018.

\bibitem{arena2022overview}
F.~Arena, M.~Collotta, G.~Pau, and F.~Termine, ``An overview of augmented reality,'' \emph{Computers}, vol.~11, no.~2, p.~28, 2022.

\bibitem{zhu2023egoobjects}
C.~Zhu, F.~Xiao, A.~Alvarado, Y.~Babaei, J.~Hu, H.~El-Mohri, S.~Culatana, R.~Sumbaly, and Z.~Yan, ``Egoobjects: A large-scale egocentric dataset for fine-grained object understanding,'' in \emph{Proceedings of the IEEE/CVF International Conference on Computer Vision}, 2023, pp. 20\,110--20\,120.

\bibitem{liu2022hoi4d}
Y.~Liu, Y.~Liu, C.~Jiang, K.~Lyu, W.~Wan, H.~Shen, B.~Liang, Z.~Fu, H.~Wang, and L.~Yi, ``Hoi4d: A 4d egocentric dataset for category-level human-object interaction,'' in \emph{Proceedings of the IEEE/CVF Conference on Computer Vision and Pattern Recognition}, 2022, pp. 21\,013--21\,022.

\bibitem{lu2019learning}
M.~Lu, D.~Liao, and Z.-N. Li, ``Learning spatiotemporal attention for egocentric action recognition,'' in \emph{Proceedings of the IEEE/CVF International Conference on Computer Vision Workshops}, 2019, pp. 0--0.

\bibitem{sudhakaran2019lsta}
S.~Sudhakaran, S.~Escalera, and O.~Lanz, ``Lsta: Long short-term attention for egocentric action recognition,'' in \emph{Proceedings of the IEEE/CVF Conference on Computer Vision and Pattern Recognition}, 2019, pp. 9954--9963.

\bibitem{furnari2020rolling}
A.~Furnari and G.~M. Farinella, ``Rolling-unrolling lstms for action anticipation from first-person video,'' \emph{IEEE transactions on pattern analysis and machine intelligence}, vol.~43, no.~11, pp. 4021--4036, 2020.

\bibitem{cho2022transformer}
H.~Cho and S.~Baek, ``Transformer-based action recognition in hand-object interacting scenarios,'' \emph{arXiv preprint arXiv:2210.11387}, 2022.

\bibitem{xu2023egopca}
Y.~Xu, Y.-L. Li, Z.~Huang, M.~X. Liu, C.~Lu, Y.-W. Tai, and C.-K. Tang, ``Egopca: A new framework for egocentric hand-object interaction understanding,'' in \emph{Proceedings of the IEEE/CVF International Conference on Computer Vision}, 2023, pp. 5273--5284.

\bibitem{vaswani2017attention}
A.~Vaswani, N.~Shazeer, N.~Parmar, J.~Uszkoreit, L.~Jones, A.~N. Gomez, {\L}.~Kaiser, and I.~Polosukhin, ``Attention is all you need,'' \emph{Advances in neural information processing systems}, vol.~30, 2017.

\bibitem{lin2022egocentric}
K.~Q. Lin, J.~Wang, M.~Soldan, M.~Wray, R.~Yan, E.~Z. XU, D.~Gao, R.-C. Tu, W.~Zhao, W.~Kong \emph{et~al.}, ``Egocentric video-language pretraining,'' \emph{Advances in Neural Information Processing Systems}, vol.~35, pp. 7575--7586, 2022.

\bibitem{pramanick2023egovlpv2}
S.~Pramanick, Y.~Song, S.~Nag, K.~Q. Lin, H.~Shah, M.~Z. Shou, R.~Chellappa, and P.~Zhang, ``Egovlpv2: Egocentric video-language pre-training with fusion in the backbone,'' in \emph{Proceedings of the IEEE/CVF International Conference on Computer Vision}, 2023, pp. 5285--5297.

\bibitem{radevski2023multimodal}
G.~Radevski, D.~Grujicic, M.~Blaschko, M.-F. Moens, and T.~Tuytelaars, ``Multimodal distillation for egocentric action recognition,'' in \emph{Proceedings of the IEEE/CVF International Conference on Computer Vision}, 2023, pp. 5213--5224.

\bibitem{gong2023mmg}
X.~Gong, S.~Mohan, N.~Dhingra, J.-C. Bazin, Y.~Li, Z.~Wang, and R.~Ranjan, ``Mmg-ego4d: Multimodal generalization in egocentric action recognition,'' in \emph{Proceedings of the IEEE/CVF Conference on Computer Vision and Pattern Recognition}, 2023, pp. 6481--6491.

\bibitem{chatterjee2023opening}
D.~Chatterjee, F.~Sener, S.~Ma, and A.~Yao, ``Opening the vocabulary of egocentric actions,'' \emph{arXiv preprint arXiv:2308.11488}, 2023.

\bibitem{yu2023efficient}
K.~P. Yu, Z.~Zhang, F.~Hu, and J.~Chai, ``Efficient in-context learning in vision-language models for egocentric videos,'' \emph{arXiv preprint arXiv:2311.17041}, 2023.

\bibitem{xu2023pov}
B.~Xu, S.~Zheng, and Q.~Jin, ``Pov: Prompt-oriented view-agnostic learning for egocentric hand-object interaction in the multi-view world,'' in \emph{Proceedings of the 31st ACM International Conference on Multimedia}, 2023, pp. 2807--2816.

\bibitem{xu2023towards}
L.~Xu, Q.~Wu, L.~Pan, F.~Meng, H.~Li, C.~He, H.~Wang, S.~Cheng, and Y.~Dai, ``Towards continual egocentric activity recognition: A multi-modal egocentric activity dataset for continual learning,'' \emph{arXiv preprint arXiv:2301.10931}, 2023.

\bibitem{kurita2023refego}
S.~Kurita, N.~Katsura, and E.~Onami, ``Refego: Referring expression comprehension dataset from first-person perception of ego4d,'' in \emph{Proceedings of the IEEE/CVF International Conference on Computer Vision}, 2023, pp. 15\,214--15\,224.

\bibitem{wang2023ego}
H.~Wang, M.~K. Singh, and L.~Torresani, ``Ego-only: Egocentric action detection without exocentric transferring,'' in \emph{Proceedings of the IEEE/CVF International Conference on Computer Vision}, 2023, pp. 5250--5261.

\bibitem{huang2023egocentric}
C.~Huang, Y.~Tian, A.~Kumar, and C.~Xu, ``Egocentric audio-visual object localization,'' in \emph{Proceedings of the IEEE/CVF Conference on Computer Vision and Pattern Recognition}, 2023, pp. 22\,910--22\,921.

\bibitem{wu2023localizing}
T.-L. Wu, Y.~Zhou, and N.~Peng, ``Localizing active objects from egocentric vision with symbolic world knowledge,'' \emph{arXiv preprint arXiv:2310.15066}, 2023.

\bibitem{gonzalez2022real}
E.~Gonzalez-Sosa, A.~Gajic, D.~Gonzalez-Morin, G.~Robledo, P.~Perez, and A.~Villegas, ``Real time egocentric segmentation for video-self avatar in mixed reality,'' \emph{arXiv preprint arXiv:2207.01296}, 2022.

\bibitem{yemisi2023fine}
E.~Yemisi-Babatope, M.~{\'A}. Camargo-Rojas, M.~S. Garc{\'\i}a-V{\'a}zquez, and A.~A. Ram{\'\i}rez-Acosta, ``Fine-tuned deep convolutional neural network for hand segmentation in egocentric videos,'' in \emph{Optics and Photonics for Information Processing XVII}, vol. 12673.\hskip 1em plus 0.5em minus 0.4em\relax SPIE, 2023, pp. 15--23.

\bibitem{chen2017deeplab}
L.-C. Chen, G.~Papandreou, I.~Kokkinos, K.~Murphy, and A.~L. Yuille, ``Deeplab: Semantic image segmentation with deep convolutional nets, atrous convolution, and fully connected crfs,'' \emph{IEEE transactions on pattern analysis and machine intelligence}, vol.~40, no.~4, pp. 834--848, 2017.

\bibitem{lu2021simpler}
Z.~Lu, S.~He, X.~Zhu, L.~Zhang, Y.-Z. Song, and T.~Xiang, ``Simpler is better: Few-shot semantic segmentation with classifier weight transformer,'' in \emph{Proceedings of the IEEE/CVF International Conference on Computer Vision}, 2021, pp. 8741--8750.

\bibitem{liu2020part}
Y.~Liu, X.~Zhang, S.~Zhang, and X.~He, ``Part-aware prototype network for few-shot semantic segmentation,'' in \emph{Computer Vision--ECCV 2020: 16th European Conference, Glasgow, UK, August 23--28, 2020, Proceedings, Part IX 16}.\hskip 1em plus 0.5em minus 0.4em\relax Springer, 2020, pp. 142--158.

\bibitem{yang2020prototype}
B.~Yang, C.~Liu, B.~Li, J.~Jiao, and Q.~Ye, ``Prototype mixture models for few-shot semantic segmentation,'' in \emph{Computer Vision--ECCV 2020: 16th European Conference, Glasgow, UK, August 23--28, 2020, Proceedings, Part VIII 16}.\hskip 1em plus 0.5em minus 0.4em\relax Springer, 2020, pp. 763--778.

\bibitem{bucher2019zero}
M.~Bucher, T.-H. Vu, M.~Cord, and P.~P{\'e}rez, ``Zero-shot semantic segmentation,'' \emph{Advances in Neural Information Processing Systems}, vol.~32, 2019.

\bibitem{li2020consistent}
P.~Li, Y.~Wei, and Y.~Yang, ``Consistent structural relation learning for zero-shot segmentation,'' \emph{Advances in Neural Information Processing Systems}, vol.~33, pp. 10\,317--10\,327, 2020.

\bibitem{ding2022decoupling}
J.~Ding, N.~Xue, G.-S. Xia, and D.~Dai, ``Decoupling zero-shot semantic segmentation,'' in \emph{Proceedings of the IEEE/CVF Conference on Computer Vision and Pattern Recognition}, 2022, pp. 11\,583--11\,592.

\bibitem{huang2018weakly}
Z.~Huang, X.~Wang, J.~Wang, W.~Liu, and J.~Wang, ``Weakly-supervised semantic segmentation network with deep seeded region growing,'' in \emph{Proceedings of the IEEE conference on computer vision and pattern recognition}, 2018, pp. 7014--7023.

\bibitem{he2016deep}
K.~He, X.~Zhang, S.~Ren, and J.~Sun, ``Deep residual learning for image recognition,'' in \emph{Proceedings of the IEEE conference on computer vision and pattern recognition}, 2016, pp. 770--778.

\bibitem{li2021pseudo}
Y.~Li, Z.~Kuang, L.~Liu, Y.~Chen, and W.~Zhang, ``Pseudo-mask matters in weakly-supervised semantic segmentation,'' in \emph{Proceedings of the IEEE/CVF international conference on computer vision}, 2021, pp. 6964--6973.

\bibitem{lee2021anti}
J.~Lee, E.~Kim, and S.~Yoon, ``Anti-adversarially manipulated attributions for weakly and semi-supervised semantic segmentation,'' in \emph{Proceedings of the IEEE/CVF Conference on Computer Vision and Pattern Recognition}, 2021, pp. 4071--4080.

\bibitem{lee2022threshold}
M.~Lee, D.~Kim, and H.~Shim, ``Threshold matters in wsss: Manipulating the activation for the robust and accurate segmentation model against thresholds,'' in \emph{Proceedings of the IEEE/CVF Conference on Computer Vision and Pattern Recognition}, 2022, pp. 4330--4339.

\bibitem{su2021context}
Y.~Su, R.~Sun, G.~Lin, and Q.~Wu, ``Context decoupling augmentation for weakly supervised semantic segmentation,'' in \emph{Proceedings of the IEEE/CVF international conference on computer vision}, 2021, pp. 7004--7014.

\bibitem{zhang2020reliability}
B.~Zhang, J.~Xiao, Y.~Wei, M.~Sun, and K.~Huang, ``Reliability does matter: An end-to-end weakly supervised semantic segmentation approach,'' in \emph{Proceedings of the AAAI Conference on Artificial Intelligence}, vol.~34, no.~07, 2020, pp. 12\,765--12\,772.

\bibitem{zhang2021complementary}
F.~Zhang, C.~Gu, C.~Zhang, and Y.~Dai, ``Complementary patch for weakly supervised semantic segmentation,'' in \emph{Proceedings of the IEEE/CVF international conference on computer vision}, 2021, pp. 7242--7251.

\bibitem{zhang2021adaptive}
X.~Zhang, Z.~Peng, P.~Zhu, T.~Zhang, C.~Li, H.~Zhou, and L.~Jiao, ``Adaptive affinity loss and erroneous pseudo-label refinement for weakly supervised semantic segmentation,'' in \emph{Proceedings of the 29th ACM International Conference on Multimedia}, 2021, pp. 5463--5472.

\bibitem{wang2020weakly}
X.~Wang, S.~Liu, H.~Ma, and M.-H. Yang, ``Weakly-supervised semantic segmentation by iterative affinity learning,'' \emph{International Journal of Computer Vision}, vol. 128, pp. 1736--1749, 2020.

\bibitem{lee2021railroad}
S.~Lee, M.~Lee, J.~Lee, and H.~Shim, ``Railroad is not a train: Saliency as pseudo-pixel supervision for weakly supervised semantic segmentation,'' in \emph{Proceedings of the IEEE/CVF conference on computer vision and pattern recognition}, 2021, pp. 5495--5505.

\bibitem{wang2020self}
Y.~Wang, J.~Zhang, M.~Kan, S.~Shan, and X.~Chen, ``Self-supervised equivariant attention mechanism for weakly supervised semantic segmentation,'' in \emph{Proceedings of the IEEE/CVF conference on computer vision and pattern recognition}, 2020, pp. 12\,275--12\,284.

\bibitem{yao2021non}
Y.~Yao, T.~Chen, G.-S. Xie, C.~Zhang, F.~Shen, Q.~Wu, Z.~Tang, and J.~Zhang, ``Non-salient region object mining for weakly supervised semantic segmentation,'' in \emph{Proceedings of the IEEE/CVF Conference on Computer Vision and Pattern Recognition}, 2021, pp. 2623--2632.

\bibitem{sun2020mining}
G.~Sun, W.~Wang, J.~Dai, and L.~Van~Gool, ``Mining cross-image semantics for weakly supervised semantic segmentation,'' in \emph{Computer Vision--ECCV 2020: 16th European Conference, Glasgow, UK, August 23--28, 2020, Proceedings, Part II 16}.\hskip 1em plus 0.5em minus 0.4em\relax Springer, 2020, pp. 347--365.

\bibitem{liu2021cross}
W.~Liu, X.~Kong, T.-Y. Hung, and G.~Lin, ``Cross-image region mining with region prototypical network for weakly supervised segmentation,'' \emph{IEEE Transactions on Multimedia}, 2021.

\bibitem{ru2022weakly}
L.~Ru, B.~Du, Y.~Zhan, and C.~Wu, ``Weakly-supervised semantic segmentation with visual words learning and hybrid pooling,'' \emph{International Journal of Computer Vision}, vol. 130, no.~4, pp. 1127--1144, 2022.

\bibitem{jo2023mars}
S.~Jo, I.-J. Yu, and K.~Kim, ``Mars: Model-agnostic biased object removal without additional supervision for weakly-supervised semantic segmentation,'' \emph{arXiv preprint arXiv:2304.09913}, 2023.

\bibitem{dosovitskiy2020image}
A.~Dosovitskiy, L.~Beyer, A.~Kolesnikov, D.~Weissenborn, X.~Zhai, T.~Unterthiner, M.~Dehghani, M.~Minderer, G.~Heigold, S.~Gelly \emph{et~al.}, ``An image is worth 16x16 words: Transformers for image recognition at scale,'' \emph{arXiv preprint arXiv:2010.11929}, 2020.

\bibitem{li2020content}
G.~Li, G.~Kang, W.~Liu, Y.~Wei, and Y.~Yang, ``Content-consistent matching for domain adaptive semantic segmentation,'' in \emph{European conference on computer vision}.\hskip 1em plus 0.5em minus 0.4em\relax Springer, 2020, pp. 440--456.

\bibitem{lv2020cross}
F.~Lv, T.~Liang, X.~Chen, and G.~Lin, ``Cross-domain semantic segmentation via domain-invariant interactive relation transfer,'' in \emph{Proceedings of the IEEE/CVF Conference on Computer Vision and Pattern Recognition}, 2020, pp. 4334--4343.

\bibitem{qiao2023human}
N.~Qiao, Y.~Sun, C.~Liu, L.~Xia, J.~Luo, K.~Zhang, and C.-H. Kuo, ``Human-in-the-loop video semantic segmentation auto-annotation,'' in \emph{Proceedings of the IEEE/CVF Winter Conference on Applications of Computer Vision}, 2023, pp. 5881--5891.

\bibitem{selvaraju2017grad}
R.~R. Selvaraju, M.~Cogswell, A.~Das, R.~Vedantam, D.~Parikh, and D.~Batra, ``Grad-cam: Visual explanations from deep networks via gradient-based localization,'' in \emph{Proceedings of the IEEE international conference on computer vision}, 2017, pp. 618--626.

\bibitem{Li_2021_ICCV}
Y.~Li, Z.~Kuang, L.~Liu, Y.~Chen, and W.~Zhang, ``Pseudo-mask matters in weakly-supervised semantic segmentation,'' in \emph{Proceedings of the IEEE/CVF International Conference on Computer Vision (ICCV)}, October 2021, pp. 6964--6973.

\bibitem{liu2023referring}
F.~Liu, Y.~Liu, Y.~Kong, K.~Xu, L.~Zhang, B.~Yin, G.~Hancke, and R.~Lau, ``Referring image segmentation using text supervision,'' in \emph{Proceedings of the IEEE/CVF International Conference on Computer Vision}, 2023, pp. 22\,124--22\,134.

\bibitem{chen2020simple}
T.~Chen, S.~Kornblith, M.~Norouzi, and G.~Hinton, ``A simple framework for contrastive learning of visual representations,'' in \emph{International conference on machine learning}.\hskip 1em plus 0.5em minus 0.4em\relax PMLR, 2020, pp. 1597--1607.

\bibitem{krahenbuhl2011efficient}
P.~Kr{\"a}henb{\"u}hl and V.~Koltun, ``Efficient inference in fully connected crfs with gaussian edge potentials,'' \emph{Advances in neural information processing systems}, vol.~24, 2011.

\bibitem{tang2017action}
Y.~Tang, Y.~Tian, J.~Lu, J.~Feng, and J.~Zhou, ``Action recognition in rgb-d egocentric videos,'' in \emph{2017 IEEE International Conference on Image Processing (ICIP)}.\hskip 1em plus 0.5em minus 0.4em\relax IEEE, 2017, pp. 3410--3414.

\bibitem{everingham2010pascal}
M.~Everingham, L.~Van~Gool, C.~K. Williams, J.~Winn, and A.~Zisserman, ``The pascal visual object classes (voc) challenge,'' \emph{International journal of computer vision}, vol.~88, pp. 303--338, 2010.

\bibitem{lin2014microsoft}
T.-Y. Lin, M.~Maire, S.~Belongie, J.~Hays, P.~Perona, D.~Ramanan, P.~Doll{\'a}r, and C.~L. Zitnick, ``Microsoft coco: Common objects in context,'' in \emph{Computer Vision--ECCV 2014: 13th European Conference, Zurich, Switzerland, September 6-12, 2014, Proceedings, Part V 13}.\hskip 1em plus 0.5em minus 0.4em\relax Springer, 2014, pp. 740--755.

\bibitem{chen2023adversarial}
J.~Chen, W.~Lu, Y.~Li, L.~Shen, and J.~Duan, ``Adversarial learning of object-aware activation map for weakly-supervised semantic segmentation,'' \emph{IEEE Transactions on Circuits and Systems for Video Technology}, vol.~33, no.~8, pp. 3935--3946, 2023.

\bibitem{qin2024enhanced}
Z.~Qin, Y.~Chen, G.~Zhu, E.~Zhou, Y.~Zhou, Y.~Zhou, and C.~Zhu, ``Enhanced pseudo-label generation with self-supervised training for weakly-supervised semantic segmentation,'' \emph{IEEE Transactions on Circuits and Systems for Video Technology}, 2024.

\bibitem{wang2024class}
J.~Wang, T.~Dai, X.~Zhao, {\'A}.~F. Garc{\'\i}a-Fern{\'a}ndez, E.~G. Lim, and J.~Xiao, ``Class activation map calibration for weakly supervised semantic segmentation,'' \emph{IEEE Transactions on Circuits and Systems for Video Technology}, 2024.

\bibitem{van2008visualizing}
L.~Van~der Maaten and G.~Hinton, ``Visualizing data using t-sne.'' \emph{Journal of machine learning research}, vol.~9, no.~11, 2008.

\end{thebibliography}

\begin{IEEEbiography}[{\includegraphics[width=1in,height=1.25in,clip,keepaspectratio]{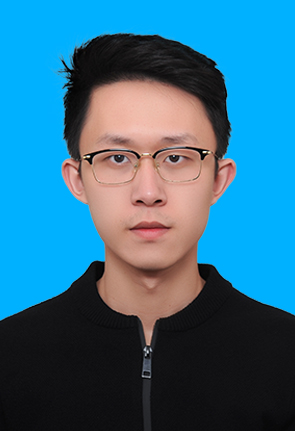}}]
{Zhaofeng Shi} received the B.E. degree in Electronic Information Engineering at the University of Electronic Science and Technology of China (UESTC) in 2021 and completed his master's studies in 2023. Now he is pursuing his Ph.D. degree in Information and Communication Engineering. His main research interests include egocentric understanding, multi-modal processing, and computer vision.
\end{IEEEbiography}

\begin{IEEEbiography}[{\includegraphics[width=1in,height=1.25in,clip,keepaspectratio]{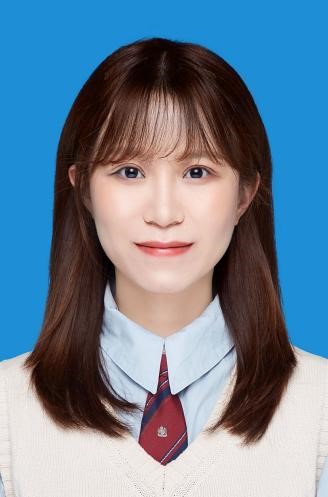}}]
{Heqian Qiu} received her Ph.D. degree in signal and information processing from the University of Electronic Science and Technology of China in 2022. She is currently a postdoctoral researcher with the School of Information and Communication Engineering, University of Electronic Science and Technology of China. She has served as a reviewer for IEEE TCSVT, JVCI, CVPR, ECCV, AAAI, etc.

Her research interests include object detection, multi-modal representative learning, computer vision, and machine learning.
\end{IEEEbiography}

\begin{IEEEbiography}[{\includegraphics[width=1in,height=1.25in,clip,keepaspectratio]{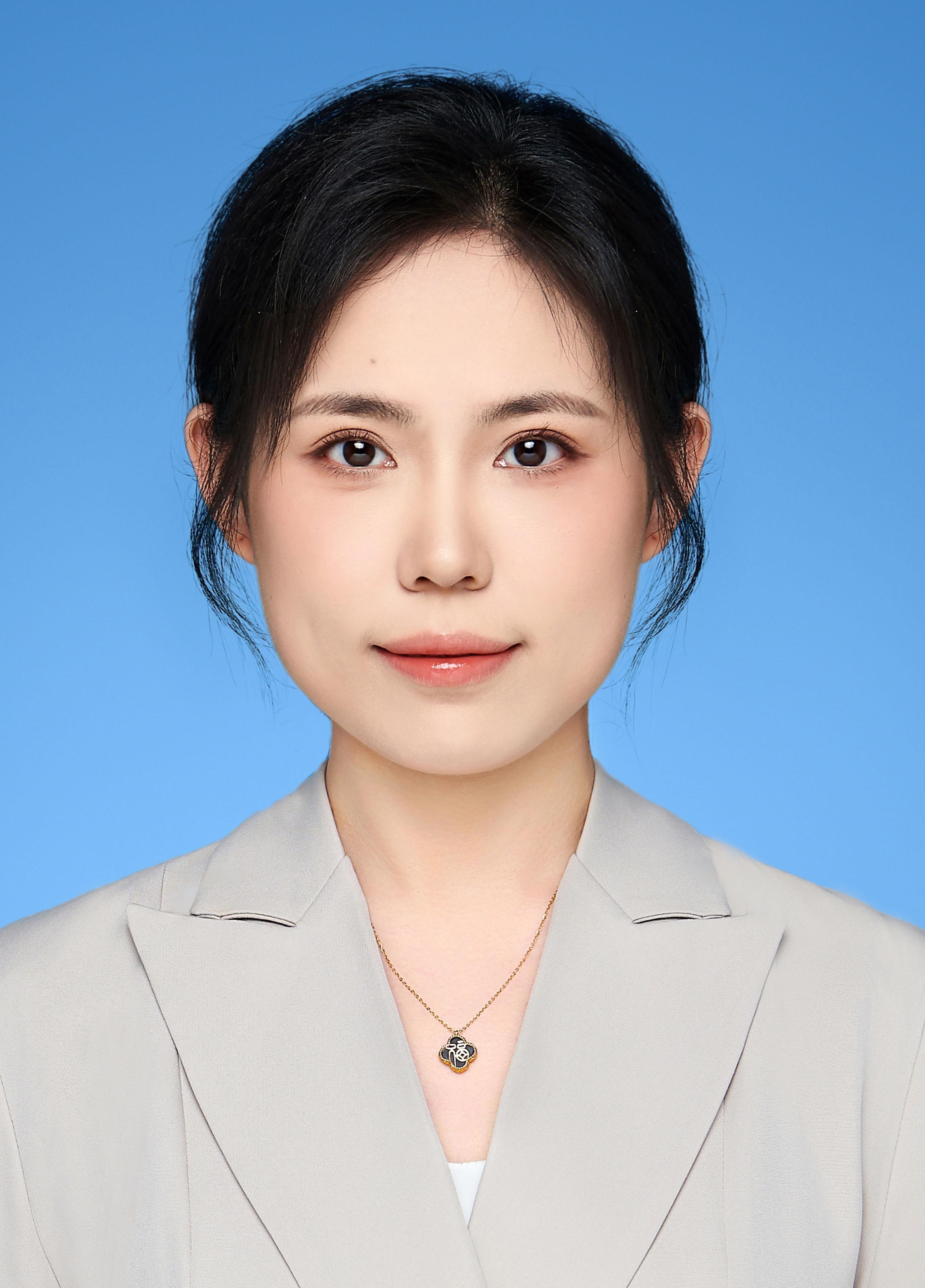}}]
{Lanxiao Wang} received her B.E. degree in Electronics Information Engineering at the University of Electronic Science and Technology of China (UESTC) in 2019. Now she is working for her Ph.D. degree under the supervision of Prof. Li. in Information and Communication Engineering at UESTC, Chengdu, China. 
	
Her main research interests include computer vision and machine learning, especially the application of deep learning on scene analysis and multimodal representation learning.
\end{IEEEbiography}

\begin{IEEEbiography}[{\includegraphics[width=1in,height=1.25in,clip,keepaspectratio]{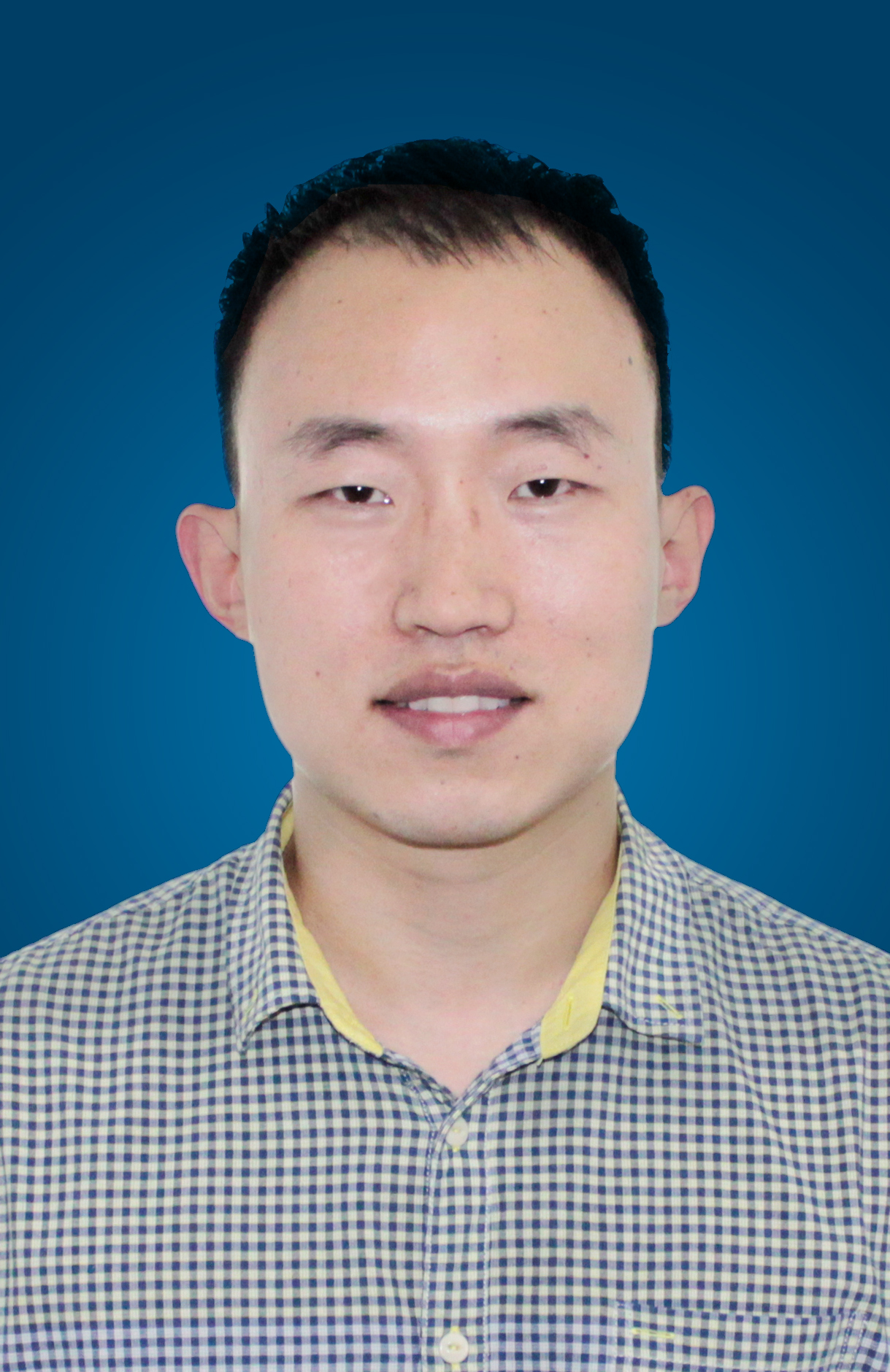}}]
{Fanman Meng} (S’12–M’14) received the Ph.D. degree in signal and information processing from the University of Electronic Science and Technology of China, Chengdu, China, in 2014. From 2013 to 2014, he was a Research Assistant with the Division of Visual and Interactive Computing, Nanyang Technological University, Singapore. He is currently Professor with the School of Information and Communication Engineering, University of Electronic Science and Technology of China. He has authored or co-authored numerous technical articles in well-known international journals and conferences. His current research interests include image segmentation and object detection. 

Dr. Meng is a member of the IEEE Circuits and Systems Society. He was a recipient of the Best Student Paper Honorable Mention Award at the 12th Asian Conference on Computer Vision, Singapore, in 2014, and the Top 10$\%$ Paper Award at the IEEE International Conference on Image Processing, Paris, France, in 2014.
\end{IEEEbiography}

\begin{IEEEbiography}[{\includegraphics[width=1in,height=1.25in,clip,keepaspectratio]{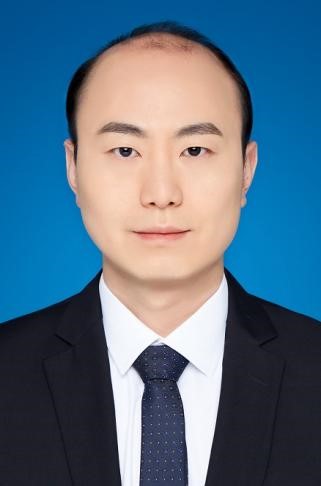}}]
{Qingbo Wu} (Member, IEEE) received the Ph.D. degree in signal and information processing from the University of Electronic Science and Technology of China in 2015. From February 2014 to May 2014, he was a Research Assistant with the Image and Video Processing (IVP) Laboratory, Chinese University of Hong Kong. From October 2014 to October 2015, he served as a Visiting Scholar with the Image and Vision Computing (IVC) Laboratory, University of Waterloo. He is currently an Associate Professor with the School of Information and Communication Engineering, University of Electronic Science and Technology of China. His research interests include image/video coding, quality evaluation, perceptual modeling and processing. He has served as Area Chair for ACM MM 2024, VCIP 2016, Session Chair for ACM MM 2021, ICMCT 2022, TPC/PC member of AAAI 2021-2023, APSIPA ASC 2020-2021, CICAI 2021-2023. He was also a Guest Editor of Remote Sensing and Frontiers in Neuroscience.
\end{IEEEbiography}

\begin{IEEEbiography}[{\includegraphics[width=1in,height=1.25in,clip,keepaspectratio]{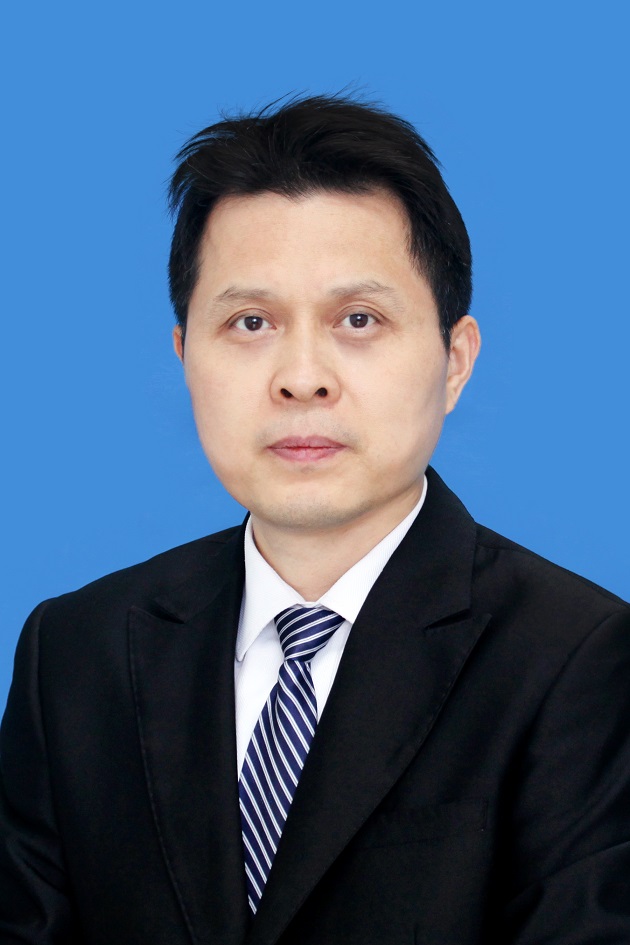}}]
{Hongliang Li} (SM’12) received his Ph.D. degree in Electronics and Information Engineering from Xi’an Jiaotong University, China, in 2005. From 2005 to 2006, he joined the visual signal processing and communication laboratory (VSPC) of the Chinese University of Hong Kong (CUHK) as a Research Associate. From 2006 to 2008, he was a Postdoctoral Fellow at the same laboratory in CUHK. He is currently a Professor in the School of Information and Communication Engineering, University of Electronic Science and Technology of China. His research interests include image and video processing, visual attention, object detection and segmentation, object recognition and parsing, multimedia content analysis, deep learning.

Dr. Li has authored or co-authored numerous technical articles in well-known international journals and conferences. He is a co-editor of a Springer book titled ``Video segmentation and its applications”. Dr. Li is involved in many professional activities. He received the 2019 and 2020 Best Associate Editor Awards for IEEE Transactions on Circuits and Systems for Video Technology (TCSVT), and the 2021 Best Editor Award for Journal on Visual Communication and Image Representation. He served as a Technical Program Chair for VCIP 2016 and PCM 2017, General Chairs for ISPACS 2017 and ISPACS 2010, a Publicity Chair for IEEE VCIP 2013, a Local Chair for the IEEE ICME 2014, Area Chairs for VCIP 2022 and 2021, and a Reviewer committee member for IEEE ISCAS from 2018 to 2022. He served as an Associate Editor of IEEE Transactions on Circuits and Systems for Video Technology (2018-2021). He is now an Associate Editor of Journal on Visual Communication and Image Representation, IEEE Open Journal of Circuits and Systems, and an Area Editor of Signal Processing: Image Communication (Elsevier Science). He is selected as the IEEE Circuits and Systems Society Distinguished Lecturer for 2022-2023.
\end{IEEEbiography}

\end{document}